\documentclass[preprint,12pt]{elsarticle}

\usepackage{lineno,hyperref}

\usepackage{epsfig}
\usepackage{graphicx}
\usepackage{amsmath}
\usepackage{amssymb}

\usepackage{xcolor}
\usepackage{soul}
\usepackage[utf8]{inputenc}
\usepackage{multirow} 
\usepackage{subfig}
\usepackage{threeparttable}
\usepackage{caption}
\usepackage{pifont}
\usepackage{booktabs}
\usepackage{algorithm}
\usepackage{algorithmic}

\newcommand{\xmark}{\ding{55}}

\begin{document}

\begin{frontmatter}

\title{Relation-Guided Representation Learning}

\author[mymainaddress,myfourthaddress]{Zhao Kang\corref{equal}}
\author[mymainaddress]{Xiao Lu\corref{equal}}%
\cortext[equal]{These authors contributed equally}
\author[mysecondaryaddress]{Jian Liang}
\author[mysecondaryaddress]{Kun Bai}
\author[myfifthaddress,mythirdaddress]{Zenglin Xu\corref{mycorrespondingauthor}}
\cortext[mycorrespondingauthor]{Corresponding author}
\address[mymainaddress]{School of Computer Science and Engineering, University of Electronic
Science and Technology of China, Sichuan, China}
\address[myfourthaddress]{Trusted Cloud Computing and Big Data Key Laboratory of Sichuan Province}
\address[mysecondaryaddress]{Cloud and Smart Industries Group, Tencent, Beijing, China}
\address[myfifthaddress]{School of Computer Science and Technology, Harbin Institute of Technology, Shenzhen, China}
\address[mythirdaddress]{Center for Artificial Intelligence, Peng Cheng Lab, Shenzhen, China}

\begin{abstract}
Deep auto-encoders (DAEs) have achieved great success in learning data representations via the powerful representability of neural networks. But most DAEs only focus on the most dominant structures which are able to reconstruct the data from a latent space and neglect rich latent structural information. In this work, we propose a new representation learning method that explicitly models and leverages sample relations, which in turn is used as supervision to guide the representation learning. Different from previous work, our framework well preserves the relations between samples. Since the prediction of pairwise relations themselves is a fundamental problem, our model adaptively learns them from data. This provides much flexibility to encode real data manifold. The important role of relation and representation learning is evaluated on the clustering task. Extensive experiments on benchmark data sets demonstrate the superiority of our approach. By seeking to embed samples into subspace, we further show that our method can address the large-scale and out-of-sample problem. Our source code is publicly available at: \url{https://github.com/nbShawnLu/RGRL}.
\end{abstract}


\end{frontmatter}


\section{Introduction}
Acquiring useful representations is crucial to the performance of numerous techniques in a wide range of fields, such as machine learning, computer vision, pattern recognition. Handcrafted representation based on some professional knowledge was widely used previously \cite{coates2011analysis,tang2018robust,dalal2005histograms}. However, they are always limited to specific tasks or simple scenarios. Facing complex circumstances, they could severely degenerate. Therefore, learning task-friendly representations with little or no supervision has been a long-lasting yet challenging topic in artificial intelligence \cite{vincent2010stacked,tang2018robustSRMR,zhang2019joint,huang2020auto,zhu2017robust,shen2018discrete}. 

During the last decade, deep auto-encoders (DAEs) have achieved great success in unsupervised representation learning and considerable gains are obtained consequently \cite{bengio2007greedy,caron2018deep,jiang2016variational,ghasedi2017deep,jabi2019deep,QIANXuezhong}. Basically, the goal of auto-encoders is to learn a mapping of the input data to a lower-dimensional representation space which succinctly captures the statistics of an underlying data distribution \cite{bengio2013representation}. Most methods simply combine a well-designed clustering assignment loss with reconstruct loss \cite{xie2016unsupervised,guo2017improved,yang2017towards,zhu2019PR}. Some methods take pairwise relations and graphs into consideration \cite{chang2017deep,ji2017deep}. Recently, adversarial strategy has been widely used with DAEs to improve representation and clustering robustness \cite{chen2016infogan,zhou2018deep,mrabah2019adversarial}. In multi-view clustering, shared generative latent representation learning \cite{yin2019shared} learns a shared latent representation under the VAE framework. AE$^2$-Nets \cite{zhang2019ae2} jointly learns the representation of each view and encodes them into an intact latent representation with a nested auto-encoder framework. Affine Equivariant Autoencoder (AEAE) \cite{guo2019affine} learns features that are equivariant to the affine transformation.

Though impressive performance has been achieved, some important structural information, e.g., pairwise relation, is not well taken care of \cite{wang2017feature}. Pairwise relations, i.e., similarities, between data samples play an important role in many applications of artificial intelligence \cite{hofmann1997pairwise,Tang2018MVUFS,scholkopf2001learning,tang2019unsupervised}. Many traditional dimensionality reduction methods such as kernel PCA \cite{scholkopf1998nonlinear,peng2020robust}, isomap \cite{tenenbaum2000global}, t-SNE \cite{maaten2008visualizing}, matrix factorization \cite{huang2020auto}, and locally linear embedding (LLE) \cite{tenenbaum2000global} find low-dimensional representations of data samples by feat of retaining their pairwise relations or local neighborhoods in the embedding space. 

Besides implicit usage in dimensionality reduction task, pairwise relation is also a fundamental quantity in many other applications, e.g., $k$-nearest neighbor search, classification \cite{zhang2020twin}, clustering \cite{huang2020auto,kannan2004clusterings,TangTMM2018,shi2000normalized,zhan2017graph}, kernel methods \cite{ren2020consensus,scholkopf2001learning,ren2020simultaneous}. In particular, the performance of spectral clustering \cite{kang2019partition,ng2002spectral} heavily depends on the quality of the input similarity graph matrix. The prediction of pairwise relations themselves is at the heart of these methods \cite{kang2019cyber,zhang2018generalized}. Pre-defined pairwise relation is rather heuristic and might not be able to reflect the intrinsic data structure. It will be highly desirable to be able to automatically learn pairwise relation that would work the underlying data \cite{kang2020structure}.

\begin{figure}[!htp]
\centering
{\includegraphics[width=.9\textwidth]{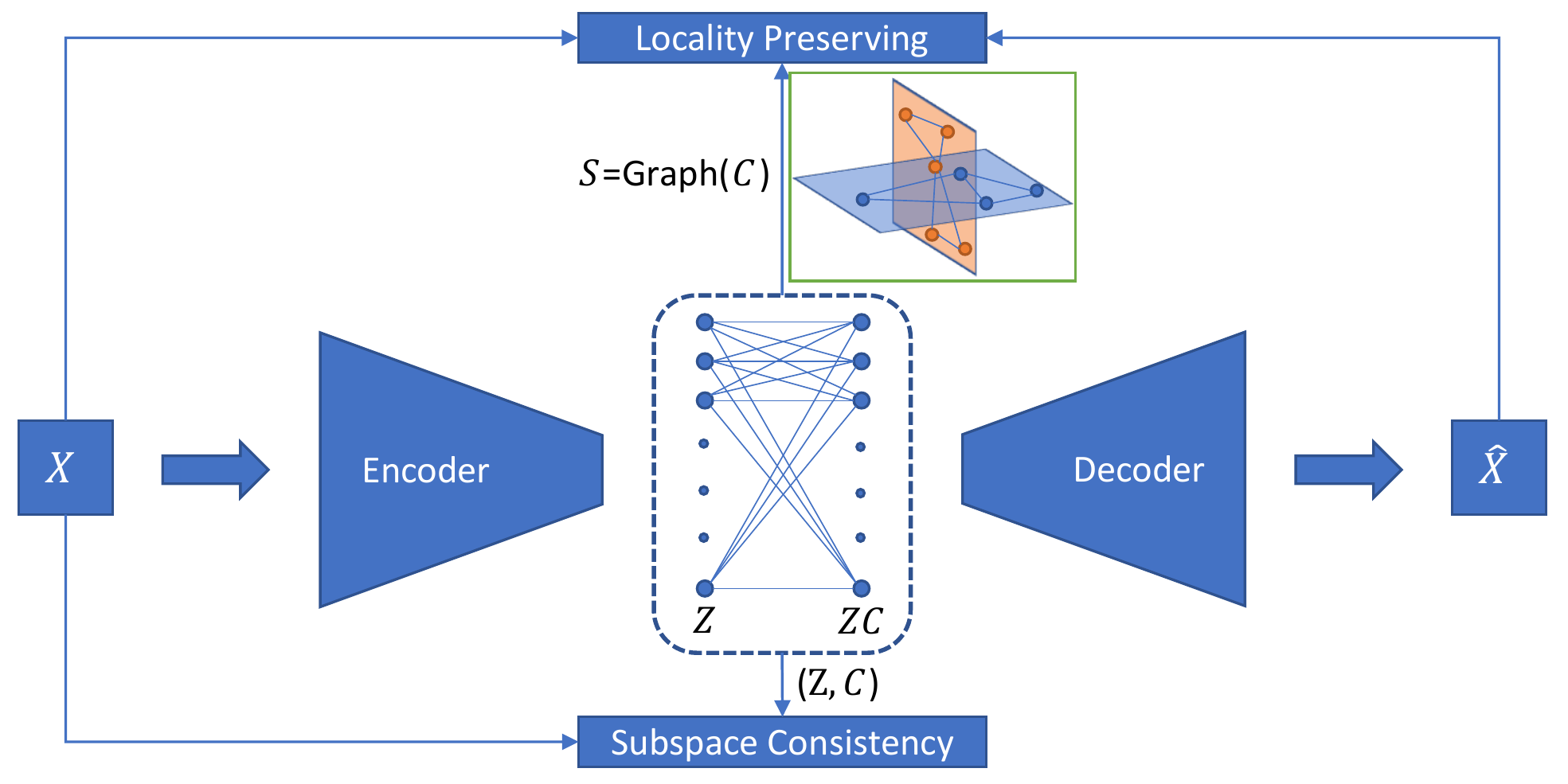}}
\caption{Architecture of RGRL. The input $X$ is mapped to $Z$ through an encoder, $Z$ is self-expressed by $ZC$, and then reconstructed as $\hat{X}$ through a decoder. Pairwise relation $C$ guides the learning process of low-dimensional representation. Instead of self-reconstruction in conventional auto-encoder, weighted reconstruction is applied to preserve locality structure information. Subspace consistency is harnessed to ensure the cluster structure is not hurt after the transformation. }
\label{arch}

\end{figure} 
In this paper, we explore a data-driven approach to learn pairwise relation and low-dimensional representation simultaneously, named Relation-Guided Representation Learning (RGRL). RGRL takes advantage of relations among samples. Specifically, to discover the underlying manifold structure and obtain a more informative representation, we don't adopt the widely used reconstruction loss in auto-encoder, i.e., each instance is reconstructed by itself, which fails to explicitly model the data relation. In our framework, each instance $x_i$ is reconstructed by a set of instances $\hat{x}_j$ weighted by the corresponding relation between them, so as to satisfy locality preserving. Rather than using fixed relations, we iteratively learn them from data. This learning process is based on the self-expression property, i.e., each sample can be represented by a linear combination of other samples in the same subspace. Moreover, this relation should hold for both raw data and embeddings, i.e., subspace consistency.

The framework of our method is displayed in Figure \ref{arch}. The weights of the self-expression layer correspond to the combination coefficient $C$, i.e., the relations between data instances. Locality preserving ensures the reconstruction of data, while subspace consistency guarantees pairwise relations hold before and after transformation. Although the fundamental idea applies to data in many tasks, we focus on clustering in this paper.

In summary, our contributions can be summarized as:
\begin{itemize}
    \item {We explicitly model and leverage relations between samples to guide the representation learning for deep auto-encoder. The embedded representations well preserve the local neighborhood structure on the data manifold.}
    \item {The proposed method learns pairwise relation or similarity and maintains its consistency in both input space and embedding space. Consequently, the subspace or cluster structure is retained. }
    \item {Extensive experiments on clustering, similarity learning, and embedding demonstrate the superiority of our method. In particular, we show how to tackle large-scale data challenge based on out-of-sample approach.}
\end{itemize}

The paper is organized as follows: Section \ref{secrelated} gives a brief review about related works. Section \ref{secmethod} introduces locality preserving and subspace consistency, then we propose our relation-guided representation learning method. In Section \ref{secsim-exp}, we implement our method on clustering task and compare with some related methods. In Section \ref{secemb-exp}, we extend our method for large-scale datasets and examine embedding performance with experiments. The paper is concluded in Section \ref{secconclusion}. 

\section{Related Work}
\label{secrelated}
In this paper, we concentrate on relation preserving embedding construction, similarity learning, and its application to clustering. Thus, we give a brief review of some related work. 

Deep auto-encoder aims to compress data $X\in\mathcal{R}^{d\times n}$ into low-dimensional representation $Z\in\mathcal{R}^{k\times n}$ where $k\ll d$, which in turn reconstructs the original data. It is often composed of a encoder $F$ and a decoder $G$ with mirror construction. Their parameters are denoted as $\Theta_e$ and $\Theta_d$, respectively. Specifically, an auto-encoder can be optimized by objective function
\begin{equation}
 \min_{\Theta_e,\Theta_d} \sum\limits_{i=1}^{n} \|X_i-G_{\Theta_d}(F_{\Theta_e}(X_i))\|^2.\label{ae}
\end{equation}

Though this basic model has led to far-reaching success for data representation, it forces to reconstruct its input without considering other data points present in the data \cite{wang2017feature}. To overcome this limitation, \cite{wang2014generalized} proposes a generalized auto-encoder framework to capture the local manifold structure. The relation is pre-calculated based on some heuristic functions, e.g., Cosine, Gaussian. This approach has one inherent limitation, i.e., it might not be appropriate to the structure of the data space \cite{shakhnarovich2005learning}. In deep manifold clustering (DMC) \cite{chen2017unsupervised}, the authors interpret the locality of manifold as similar inputs should have similar representations, and minimize the reconstruction of $X_i$ itself and its local neighborhood. However, they define reconstruction weights either in a supervised or pre-defined way and the relations between samples are not flexible for modeling. 

To keep locality properties on latent space is another way to get effective representations. The deep embedding network (DEN) \cite{huang2014deep} first learns representations from an auto-encoder while keeps locality-preserving and group sparsity constraints on low-dimension space. It requires that two latent representations should be similar if they are similar in the original space defined by Gaussian kernel. The deep embedded clustering (DEC) \cite{xie2016unsupervised} method fine-tunes the encoder by minimizing KL divergence between soft assignment and target distribution. The improved deep embedded clustering (IDEC) \cite{guo2017improved} improves DEC by remaining decoder in fine-tune stage. The recently developed deep $k$-means (DKM) \cite{fard2018deep} jointly learns latent representations and $k$-means. 

Motivated by the success of subspace clustering method, deep subspace clustering (DSC) \cite{ji2017deep} implements subspace clustering method in deep neural network and achieves promising results. However, DSC only assumes the subspace structure in latent space, which fails to make full use of the original data, and forces to reconstruct all parts of the input, even if they are contaminated by noise or outliers. \cite{kheirandishfard2020multi} points out that different layers of the encoder provide different information and it is difficult to find a suitable subspace clustering representation by only relying on the output of the encoder. This provides us a strong motivation to incorporate original information to enhance the clustering performance. The deep adversarial subspace clustering (DASC) \cite{zhou2018deep} improves DSC by introducing adversarial learning so that the discriminator can evaluate the clustering quality and supervise the generator’s learning. \cite{huang2019deep} addresses the outliers and initialization issues by adding a weighted subspace network.

Recently, self-supervised learning becomes a popular tool of unsupervised learning which uses pretext tasks to replace the labels annotated by humans\cite{doersch2015unsupervised,pathak2017learning}. The DeepCluster \cite{caron2018deep} uses the cluster assignments as pseudo-labels to learn the parameters of the
network. The deep comprehensive correlation mining (DCCM) \cite{wu2019deep} makes use of the local robustness assumption and utilizes above pseudo-graph and pseudo-label to learn better representation. The self-supervised convolutional subspace clustering (S$^2$ConvSCN) \cite{zhang2019self} introduces a spectral clustering module and a classification module into DSC, i.e., applying the current clustering results to supervise the training of network. All deep subspace clustering methods have large cost of memory due to the self-expression layer  structure,  which  hinders  their  applications  on  large-scale data  sets.  This  explains  why  all  DSC  methods  use  small  datasets.

In this paper, we aim to learn representations guided by locality preserving and global subspace consistency using a simple and neat model. We compare our method with some of the related work in Table \ref{approach}. We can clearly see the advantage of our approach. Meanwhile, more complicated modules and tricks can be easily apply on the top of our fundamental model.

\begin{table}
\centering
\caption{Comparison of recently proposed unsupervised learning methods with our approach. CNN denotes using convolutional neural network, SL denotes subspace learning, LP denotes the locality preserving, SC denotes subspace consistency.}
\renewcommand{\arraystretch}{1.}
\resizebox{.8\textwidth}{!}{
\begin{tabular}{c|c c c c c c c c c}
\hline
Approach & DMC & DEC & IDEC & DKM & DSC & DEPICT & DSCDAN & RGRL \\
\hline
CNN & \xmark  & \xmark & \xmark & \xmark & \checkmark & \xmark & \xmark & \checkmark  \\
\hline
SL & \xmark  & \xmark & \xmark & \xmark & \checkmark & \xmark & \xmark & \checkmark \\
\hline
LP & \checkmark  & \xmark & \checkmark & \xmark & \xmark &\checkmark & \checkmark & \checkmark \\
\hline
SC & \xmark  & \xmark & \xmark & \xmark & \xmark & \xmark & \xmark & \checkmark \\
\hline
\end{tabular}}
\label{approach}
\end{table}

\section{The Proposed Method}
\label{secmethod}
We propose a deep auto-encoder network to learn representations of the data guided by the relation between samples. The proposed approach is composed of four key components: 1) the encoder $F$ encodes $X$ into latent representation $Z$, 2) the decoder $G$ reconstructs $\hat{X}$ from latent representation, 3) the locality preserving module, 4) the subspace consistency module. Our goal is to train a reconstruction such that $\hat{X_i}$ is not only similar to $X_i$, but also to other samples $X_j$ determined by their relations $C_{ij}$. Notably, our used network architecture is similar to DSC \cite{ji2017deep}. Different from DSC, our main contribution lies in the design of two different objective terms, i.e., locality preserving and subspace consistency, aiming to fully exploit the relations among data points. By contrast, DSC applies traditional self-reconstruction loss and self-expression only in latent space.
\subsection{Locality Preserving}
To preserve the local structure, we use the weighted reconstruction instead of Eq.(\ref{ae}). To be precise, $X_i$ is reconstructed by $\hat{X}_j$ with weight $S_{ij}$, where $S_{ij}$ is the similarity between samples $X_i$ and $X_j$. Samples with large distance should have low similarity. Then, the network can be trained by solving
\begin{equation}
\min_{\Theta_e,\Theta_d}{\sum_{ij} S_{ij}||X_i-G_{\Theta_d}(F_{\Theta_e}(X_j))||^2}.
\label{GAE-f}
\end{equation}
Compared to self-reconstruction, Eq.(\ref{GAE-f}) can characterize the neighborhood relations of samples. In other words, the learned latent representation learned is encoded by neighborhood relations. 
Eq.(\ref{GAE-f}) can be further transformed as follows
\begin{equation}
\begin{split}
\sum S_{ij}||X_i-\hat{X}_j||^2&=\sum S_{ij}(||X_i||^2-2X_i^T\hat{X}_j+||\hat{X}_j||^2)\\
&=\sum S_{ij}[(||X_i||^2-2X_i^T\hat{X}_i+||\hat{X}_i||^2)\\
&\quad\quad\quad\quad+2(X_i^T\hat{X}_i - X_i^T\hat{X_j})]\\
&=Tr[{(X-\hat{X})}^TD(X-\hat{X})]\\
&\quad\quad\quad\quad+ 2Tr(X^TL\hat{X}),
\end{split}
\end{equation}
where diagonal matrix $D=Diag(\sum_{j=1}^nS_{ij})$ and $L=D-S$ is the Laplacian matrix. We can see that similarity matrix $S$ would be crucial to the performance of the network. Unlike many existing work using pre-defined values, we propose to automatically learn $S$ from data.

\subsection{Subspace Consistency}
Recently, similarity learning based on self-expression has been widely used. Its basic idea is that each sample can be represented by a linear combination of other samples in the same subspace \cite{elhamifar2013sparse,liu2012robust,kang2017kernel}. This combination coefficient represents the relation between samples. In general, it solves the following problem
\begin{equation}
\min_C \frac{1}{2}\|X-XC\|_F^2+\alpha \|C\|_p \quad s.t.\quad diag(C)=0,
\end{equation} 
where the first term minimizes the reconstruction error, the second term is certain regularizer function on $C$, and $\alpha$ is a balance parameter. Matrix $C$ can represent the subspace structure of data, i.e., $C_{ij}=0$ if the $i$-th sample and $j$-th sample do not lie in the same subspace. 

In our case, we have two representations, i.e., original space $X$ and latent space $Z_{\Theta_e}$. We expect the subspace structure can be well preserved after the transformation, i.e., subspace consistency. Therefore, we also minimize the self-expression error in latent space. DSC \cite{ji2017deep} fails to consider the subspace structure in the raw space. Then, our objection function becomes

\begin{equation}
\begin{split}
\min_C &\frac{1}{2}\|X-XC\|_F^2+\frac{\beta}{2}\|Z_{\Theta_e}-Z_{\Theta_e}C\|_F^2+\alpha \|C\|_p\\
&\quad s.t.\quad diag(C)=0.
\end{split}
\label{LLE-f}
\end{equation} 
In the network, we add a fully connected layer without bias between the encoder and the decoder, whose weights represent the coefficient matrix $C$, so-called the self-expression layer as shown in Fig \ref{arch} \cite{ji2017deep}. By solving Eq.(\ref{LLE-f}), we can obtain the sample relation matrix $C$. Then, the similarity is usually computed based on $S = \frac{1}{2}(|C| + |C|^T)$. Since the scale of each row and column of similarity matrix might be different, we use the symmetric normalized Laplacian for scale normalization while keeping the symmetry of the similarity matrix $S$ \cite{ng2002spectral}. Specifically, normalized degree matrix $D_n=I$ and normalized Laplacian matrix $L_n=D^{-\frac{1}{2}}LD^{-\frac{1}{2}}$. Then, we have $Tr[{(X-\hat{X})}^TD_n(X-\hat{X})]$ = $\|X-\hat{X}\|_F^2$.

\subsection{Proposed Formulation}
To jointly train the network with relation guided by both subspace consistency and locality preserving, we combine above terms together, which yields

\begin{equation}
\begin{split}
L(\Theta)=&\|X-\hat{X}_{\Theta}\|_F^2+2Tr(X^TL_n\hat{X}_{\Theta})+\alpha||C||_p\\
&+\frac{\beta}{2}||Z_{\Theta_e}-Z_{\Theta_e}C||^2_F+\frac{\gamma}{2}\|X-XC\|_F^2\\
&\quad s.t.\quad diag(C)=0,
\end{split}
\label{objf}
\end{equation}
where $\Theta$ denotes the network parameters, which include encoder parameters $\Theta_e$, self-expression layer parameters $C$, and decoder parameters $\Theta_d$. Note that, the output $\hat{X}$ of the decoder is a function of $\{\Theta_e, C, \Theta_d\}$. In fact, all the unknowns in Eq.(\ref{objf}) are functions of network parameters. This network can be implemented by neural network frameworks and trained by back-propagation. Once the network architecture is optimized, we obtain the lower-dimensional representation $Z$ and relation matrix $C$. 

Compared to the existing work in the literature, our proposed RGRL has the following advantages:
\begin{itemize}
\item{The proposed model takes into account the data relation, which outputs relation preserving representations.}
\item{The designed architecture also performs similarity learning. This solves another fundamental problem. Moreover, we convert the similarity learning into network parameters optimization problem .}

\end{itemize}

\begin{figure*}[ht]
\centering
\subfloat[EYaleB\label{EYaleB-d}]{\includegraphics[width=.32\textwidth]{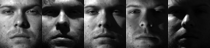}}
\hspace{0.01cm}
\subfloat[ORL\label{ORL-d}]{\includegraphics[width=.32\textwidth]{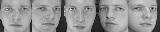}}
\hspace{0.01cm}
\subfloat[MNIST\label{MNIST-d}]{\includegraphics[width=.32\textwidth]{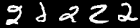}}\\
\subfloat[Umist\label{Umist-d}]{\includegraphics[width=.32\textwidth]{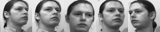}}
\hspace{0.01cm}
\subfloat[COIL20\label{COIL20-d}]{\includegraphics[width=.32\textwidth]{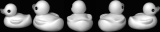}}
\hspace{0.01cm}
\subfloat[COIL40\label{COIL40-d}]{\includegraphics[width=.32\textwidth]{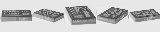}}
\caption{Examples of the datasets.}
\label{dataset-sample}
\end{figure*}

\begin{table*}
\centering
\caption{Statistics of the datasets.}
\renewcommand{\arraystretch}{1.1}
\resizebox{.9\textwidth}{!}{
\begin{tabular}{c|c c c c c c}
\hline
Dataset & EYaleB & ORL & MNIST & Umist & COIL20 & COIL40 \\
\hline
Samples & 2,432 & 400 & 1,000 & 480 & 1,440 & 2,880 \\
\hline
Classes & 38 & 40 & 10 & 20 & 20 & 40 \\
\hline
Dimensions & 48$\times$ 42 & 32$\times$ 32 & 28$\times$ 28 & 32$\times$ 32 & 32$\times$ 32 & 32$\times$ 32\\
\hline
\end{tabular}}
\label{dataset}
\end{table*}

\begin{table*}[htbp]
\centering
\caption{Network settings for clustering experiments, including the ``kernel size@channels" and size of $C$.}

\renewcommand{\arraystretch}{1.1}
\resizebox{.9\textwidth}{!}{
\begin{tabular}{c|c|c|c|c|c|c}
\hline
  & EYaleB & ORL & MNIST & Umist & COIL20 & COIL40\\
\hline 
\multirow{3}{4em}{encoder}&5$\times$5@10&5$\times$5@5&5$\times$5@15&5$\times$5@20&3$\times$3@15&3$\times$3@20\\
&3$\times $3@20&3$\times$3@3&3$\times $3@10&3$\times$3@10&-&-\\
&3$\times $3@30&3$\times $3@3&3$\times $3@5&3$\times$3@5&-&-\\
\hline
C&2432$\times$2432 &400$\times$400&1000$\times$1000&480$\times$480&1440$\times$1440&2880$\times$2880\\
\hline
\multirow{3}{4em}{decoder}&3$\times$3@30&3$\times$3@3&3$\times$3@5&3$\times$3@5&3$\times$3@15&3$\times$3@20\\
&3$\times $3@20&3$\times $3@3&3$\times $3@10&3$\times$3@10&-&-\\
&5$\times $5@10&5$\times $5@5&5$\times $5@15&5$\times$5@20&-&-\\
\hline

\end{tabular}}
\label{setup}
\end{table*}

\section{Similarity Learning Experiments}
\label{secsim-exp}
In this section, we evaluate the similarity learning effect on the clustering task. 
\subsection{Datasets}
We perform experiments on three widely used face datasets: ORL, Extended Yale B (EYaleB), Umist; and three object datasets: MNIST, COIL20, and COIL40. ORL is composed of 40 subjects, each subject has 10 images taken with varying poses and expressions. EYaleB contains 38 subjects each with 64 images taken under different illumination. Umist only contains 20 individuals, each person has 24 images taken under very different poses. COIL20 has 20 classes of toys with 72 images in each class. COIL40 has 40 classes of objects with 72 images in each class. For MNIST, we use the first 100 images of each digit. The statistics of the datasets are summarized in Table \ref{dataset}. Some examples of the datasets are shown in Fig \ref{dataset-sample}.

\subsection{Experimental Setup}

In this experiment, we use convolutional neural networks with ReLU activation function to implement the encoder and decoder. We use one layer convolutional network for COIL20 and COIL40, and three layers convolutional network for others. The architecture details of the networks are shown in Table \ref{setup}.

We first pre-train the encoder and decoder without the self-expression layer. Then we fine-tune the whole network. We fix regularization parameter $\alpha$ as 1e-4 and perform grid searching for $\beta$ and $\gamma$. We use Adam \cite{kingma2014adam} as the optimizer. Learning rate is set as 1e-3 in pre-training, and 1e-4 in fine-tune stage. Our method is implemented with Tensorflow and the experiments are run on a server with an NVIDIA TITAN Xp GPU, 12GB GRAM. 

We implement a spectral clustering algorithm after we obtain the weight $C$. Spectral clustering \cite{ng2002spectral} is a popular clustering technique with promising performance. Nevertheless, it is always challenging to construct an appropriate similarity graph that is most suitable for the specific dataset at hand. To examine the performance of learned coefficient matrix $C$, we use the coefficient matrix to build an affinity matrix $A$, as input to the spectral clustering algorithm. To enhance the block-structure and improve the clustering accuracy, we employ the approach proposed in efficient dense subspace clustering (EDSC) \cite{ji2014efficient}, which can be summarized as Algorithm \ref{alg:affinity}, where $\alpha$ is empirically selected according to the level of noise and $d$ is the maximal intrinsic dimension of subspaces. For fairness, we use the same setting as deep subspace clustering network (DSC) \cite{ji2017deep}.

\begin{algorithm}
\caption{Compute affinity matrix.} 
\label{alg:affinity} 
\begin{algorithmic}[1]
\REQUIRE
The relation matrix, $C$;\\
The number of cluster, $k$;\\
The intrinsic dimension of subspaces, $d$;
\ENSURE
The affinity matrix, $A$;  
\STATE Let $S = \frac{1}{2}(|C| + |C|^T)$;
\STATE Compute the SVD of $S$, $S = U\Sigma V^T$;
\STATE Let $Z = U_m\Sigma_m^{\frac{1}{2}}$, where $m=k*d+1$;
\STATE Compute affinity matrix $A = [ZZ^T]^\alpha$;
\end{algorithmic}
\end{algorithm}

We compare with closely related shallow and deep techniques developed in recent years. They include: low rank representation (LRR) \cite{liu2013robust}, low rank subspace clustering (LRSC) \cite{vidal2014low}, sparse subspace clustering (SSC) \cite{elhamifar2013sparse}, kernel sparse subspace clustering (KSSC) \cite{patel2014kernel}, SSC by orthogonal matching pursuit (SSC-OMP) \cite{you2016scalable}, efficient dense subspace clustering (EDSC) \cite{ji2014efficient}, SSC with pre-trained convolutional auto-encoder features (AE+SSC), deep subspace clustering network with $\ell1$-norm (DSC-L1) \cite{ji2017deep}, deep subspace clustering network with $\ell2$-norm (DSC-L2), deep embedding clustering (DEC) \cite{xie2016unsupervised}, deep $k$-means (DKM) \cite{fard2018deep}, deep comprehensive correlation mining (DCCM) \cite{wu2019deep}, deep embedded regularized clustering (DEPICT) \cite{ghasedi2017deep}, and deep spectral clustering using dual autoencoder network (DSCDAN) \cite{yang2019deep}.

\begin{table*}[!ht]
\centering
\caption{Clustering results of RGRL and compared methods on MNIST, EYaleB, ORL, COIL20, COIL40, and Umist. For the sake of space, we only list results of RGRL$_{sc}$ with $\ell2$-norm.}
\renewcommand{\arraystretch}{1.3}
\resizebox{.99\textwidth}{!}{
\begin{tabular}{c |c|c c c c c c c|c c c c c c c c c c c}
\hline
Dataset & Metric & SSC & ENSC & KSSC & SSC-OMP & EDSC & LRR & LRSC & AE+SSC & DSC-L1 & DSC-L2 & DEC & DKM & DCCM & DEPICT & DSCDAN & RGRL$_{sc}$-L2 & RGRL-L1 & RGRL-L2 \\
\hline
 & ACC & 0.4530 & 0.4983 & 0.5220 & 0.3400 & 0.5650 & 0.5386 & 0.5140 & 0.4840 & 0.7280 & 0.7500 & 0.6120 & 0.5332 & 0.4020 & 0.4240 & 0.7450 & 0.7570 & {0.8130} & \textbf{0.8140}\\
MNIST & NMI & 0.4709 & 0.5495 & 0.5623 & 0.3272 & 0.5752 & 0.5632 & 0.5576 & 0.5337 & 0.7217 & 0.7319 & 0.5743 & 0.5002 & 0.3468 & 0.4236 & 0.7110 & 0.7323 & {0.7534} & \textbf{0.7552}\\
& PUR & 0.4940 & 0.5483 & 0.5810 & 0.3560 & 0.6120 & 0.5684 & 0.5550 & 0.5290 & 0.7890 & {0.7991} & 0.6320 & 0.5647 & 0.4370 & 0.3560 & 0.7480 & 0.7980 & {0.8150} & \textbf{0.8160}\\
 \hline
& ACC & 0.7354 & 0.7537 & 0.6921 & 0.7372 & 0.8814 & 0.8499 & 0.7931 & 0.7480 & 0.9681 & 0.9733 & 0.2303 & 0.1713 & 0.1176 & 0.1094 & 0.7307 & \textbf{0.9840} & {0.9757} & {0.9753}\\
EYaleB & NMI & 0.7796 & 0.7915 & 0.7359 & 0.7803 & 0.8835 & 0.8636 & 0.8264 & 0.7833 & 0.9687 & {0.9703} & 0.4258 & 0.2704 & 0.2011 & 0.1594 & 0.8808 & \textbf{0.9776} & 0.9668 & {0.9661}\\
& PUR & 0.7467 & 0.7654 & 0.7183 & 0.7542 & 0.8800 & 0.8623 & 0.8013 & 0.7597 & 0.9711 & 0.9731 & 0.2373 & 0.1738 & 0.1312 & 0.1044 & 0.7644 & \textbf{0.9840} & {0.9757} & {0.9753}\\
\hline
& ACC & 0.7425 & 0.7525 & 0.7143 & 0.7100 & 0.7038 & 0.8100 & 0.7200 & 0.7563 & 0.8550 & 0.8600 & 0.5175 & 0.4682 & 0.6250 & 0.2800 & 0.7950 & \textbf{0.8700} & 0.8650 & \textbf{0.8700}\\
ORL & NMI & 0.8459 & 0.8540 & 0.8070 & 0.7952 & 0.7799 & 0.8603 & 0.8156 & 0.8555 & 0.9023 & 0.9034 & 0.7449 & 0.7332 & 0.7906 & 0.5764 & 0.9135 & 0.9189 & 0.9169 & \textbf{0.9215}\\
& PUR & 0.7875 & 0.7950 & 0.7513 & 0.7463 & 0.7138 & 0.8225 & 0.7542 & 0.7950 & 0.8585 & 0.8625 & 0.5400 & 0.4752 & 0.5975 & 0.1450 & 0.8025 & 0.8775 & 0.8775 & \textbf{0.8850}\\
\hline
& ACC & 0.8631 & 0.8760 & 0.7087 & 0.6410 & 0.8371 & 0.8118 & 0.7416 & 0.8711 & 0.9314 & 0.9368 & 0.7215 & 0.6651 & 0.8021 & 0.8618 & 0.7868 & 0.9451 & {0.9694} & \textbf{0.9701}\\
COIL20 & NMI & 0.8892 & 0.8952 & 0.8243 & 0.7412 & 0.8828 & 0.8747 & 0.8452 & 0.8990 & 0.9353 & 0.9408& 0.8007  & 0.7971 & 0.8639 & 0.9266 & 0.9131 & 0.9607 & 0.9748 & \textbf{0.9762}\\
& PUR & 0.8747 & 0.8892 & 0.7497 & 0.6667 & 0.8585 & 0.8361 & 0.7937 & 0.8901 & 0.9306 & 0.9397 & 0.6931 & 0.6964 & 0.7889 & 0.8319 & 0.7819 & 0.9451 & {0.9694} & \textbf{0.9701}\\
 \hline
 & ACC & 0.7191 & 0.7426 & 0.6549 & 0.4431 & 0.6870 & 0.6493 & 0.6327 & 0.7391 & 0.8003 & 0.8075 & 0.4872 & 0.5812 & 0.7691 & 0.8073& 0.7385 & 0.8135 & {0.8292} & \textbf{0.8396}\\
COIL40 & NMI & 0.8212 & 0.8380 & 0.7888 & 0.6545 & 0.8139 & 0.7828 & 0.7737 & 0.8318 & 0.8852 & 0.8941 & 0.7417 & 0.7840 & 0.8890 & \textbf{0.9291} & 0.8940 & 0.9194 & {0.9246} & {0.9284}\\
& PUR & 0.7716 & 0.7924 & 0.7284 & 0.5250 & 0.7469 & 0.7109 & 0.6981 & 0.7840 & 0.8646 & \textbf{0.8740} & 0.4163 & 0.6367 & 0.7663 & 0.8191 & 0.7726 & 0.8497 & 0.8594 & 0.8594\\
 \hline
& ACC & 0.6904 & 0.6931 & 0.6531 & 0.6438 & 0.6937 & 0.6979 & 0.6729 & 0.7042 & 0.7242 & 0.7312 & 0.5521 & 0.5106 & 0.5458 & 0.4521 & 0.6937& 0.7458 & \textbf{0.8104} & \textbf{0.8104}\\
Umist & NMI & 0.7489 & 0.7569 & 0.7377 & 0.7068 & 0.7522 & 0.7630 & 0.7498 & 0.7515 & 0.7556 & 0.7662 & 0.7125 & 0.7249 & 0.7440 & 0.6329 & \textbf{0.8816} & 0.8612 & {0.8812} & {0.8812}\\
& PUR & 0.6554 & 0.6628 & 0.6256 & 0.6171 & 0.6683 & 0.6670 & 0.6562 & 0.6785 & 0.7204 & 0.7276 & 0.5917 & 0.5685 & 0.5854 & 0.4167 & 0.7167 & 0.7875 & \textbf{0.8354} & \textbf{0.8354}\\
\hline
\end{tabular}}
\label{cluster-er}
\end{table*}

Three widely used evaluation metrics are used to evaluate the performances: accuracy (ACC), normalized mutual information (NMI), and purity (PUR) \cite{liu2020efficient,peng2018integrate}. 

Accuracy is defined as:
\begin{equation}
    \text{ACC} = \frac{\sum_{i=1}^{N}{\delta(map(l_i)=y_i)}}{N},
\end{equation}
where $\delta$ is an indicator function, $l_i$ is the clustering label for $X_i$ produced by spectral clustering, $map$ transforms the clustering label $l_i$ to its group label based on Kuhn-Munkres algorithm, and $y_i$ is the ground truth label of $X_i$.

Normalized mutual information is another popular metric
used for evaluating clustering tasks. It is defined as follows:
\begin{equation}
    \text{NMI}(Y,L)=\frac{I(Y,L)}{\sqrt{H(Y)H(L)}},
\end{equation}
where $Y$ and $L$ respectively denote ground truth label and clustering label. $I$ is mutual information which measures the information gain to the true partition after knowing the clustering result, $H$ is entropy and $\sqrt{H(Y)H(L)}$ is used to normalize the mutual information.

Purity is a simple and transparent evaluation measure which is defined as:
\begin{equation}
    \text{PUR}(Y,L)=\frac{\sum_{i=1}^{k}{\max_j{|L_i\bigcap Y_j|}}}{N},
\end{equation}
where $Y$ is the set of classes and $L$ is the set of clusters, $k$ is the number of clusters, $L_i$ denotes the set of samples belongs to $i$-th cluster, $Y_j$ denotes the set of samples belongs to $j$-th class. 

To examine the strength of locality preserving by using weighted reconstruction, we made an ablation study by replacing the weighted reconstruction with original auto-encoder reconstruction, dubbed RGRL$_{sc}$, since it only has the subspace consistency effect.

\begin{figure}[!htpb]
\centering
\subfloat[MNIST-L1\label{AMNIST-L1}]{\includegraphics[width=.3\textwidth]{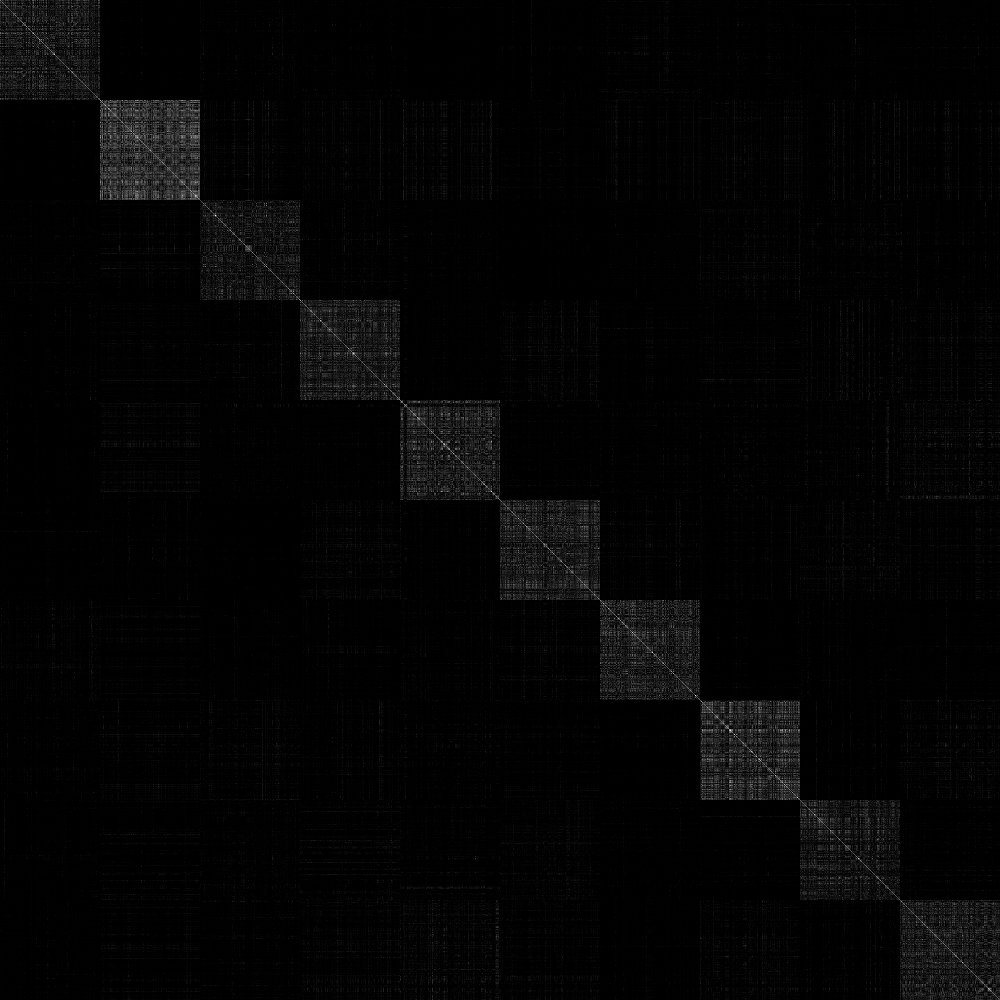}}
\hspace{.05cm}
\subfloat[MNIST-L2\label{AMNIST-L2}]{\includegraphics[width=.3\textwidth]{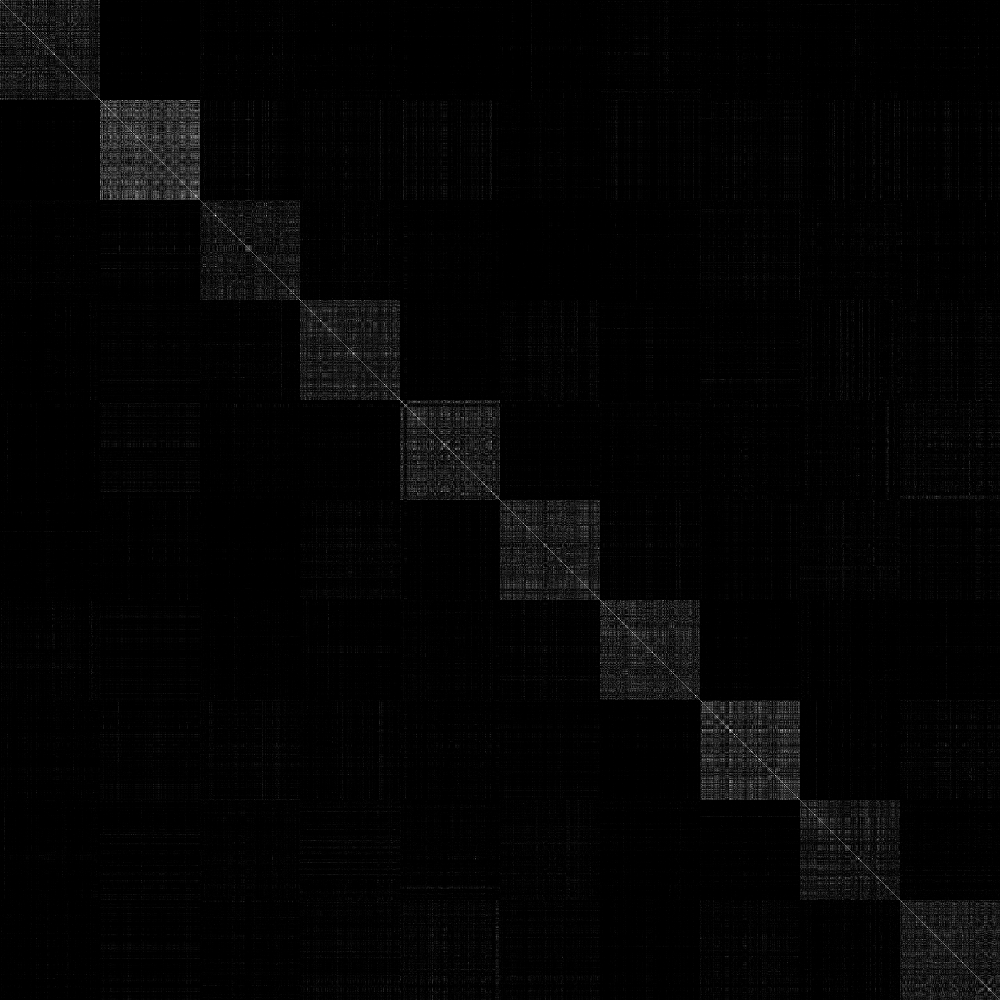}}

\subfloat[COIL20-L1\label{ACOIL20-L1}]{\includegraphics[width=.3\textwidth]{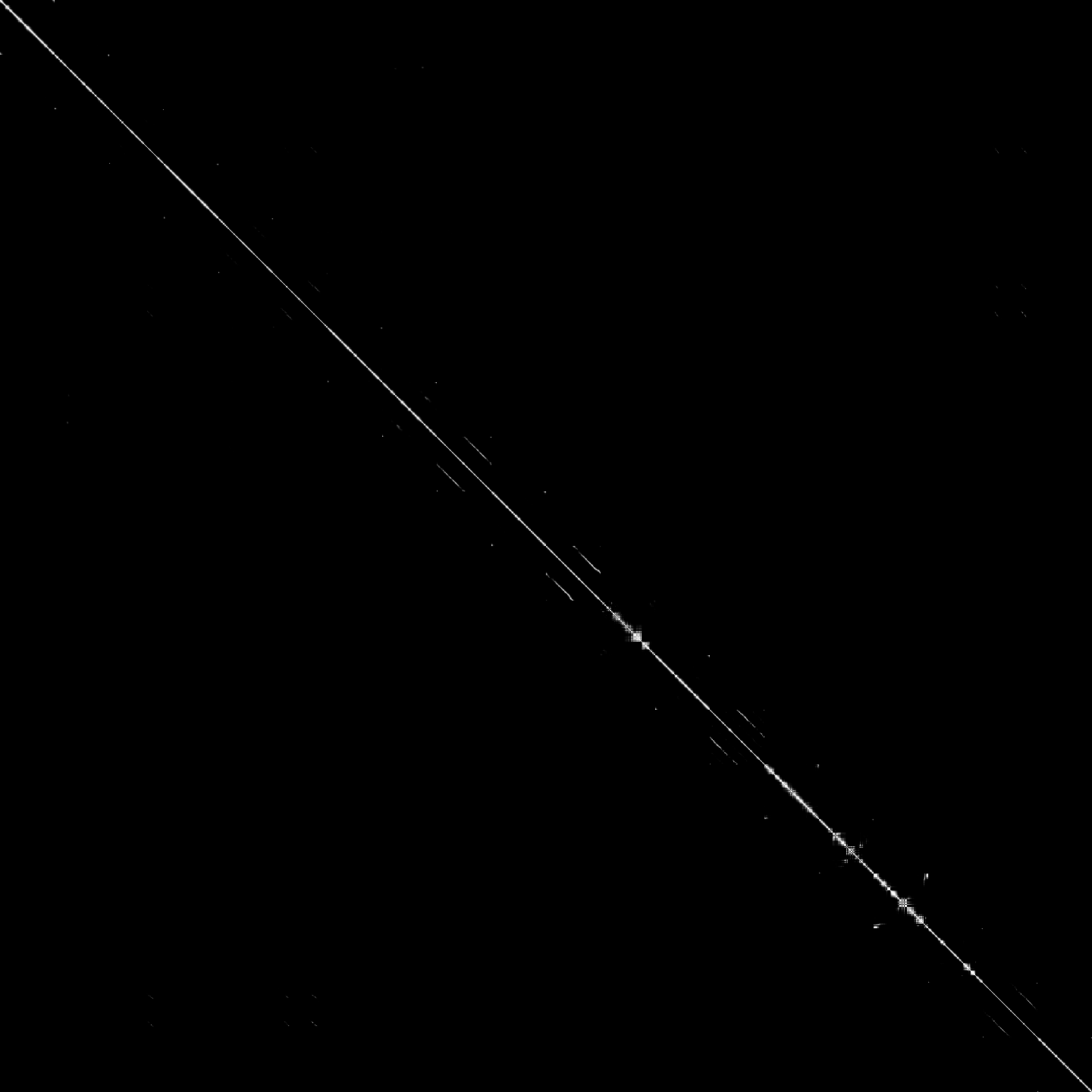}}
\hspace{.05cm}
\subfloat[COIL20-L2\label{ACOIL20-L2}]{\includegraphics[width=.3\textwidth]{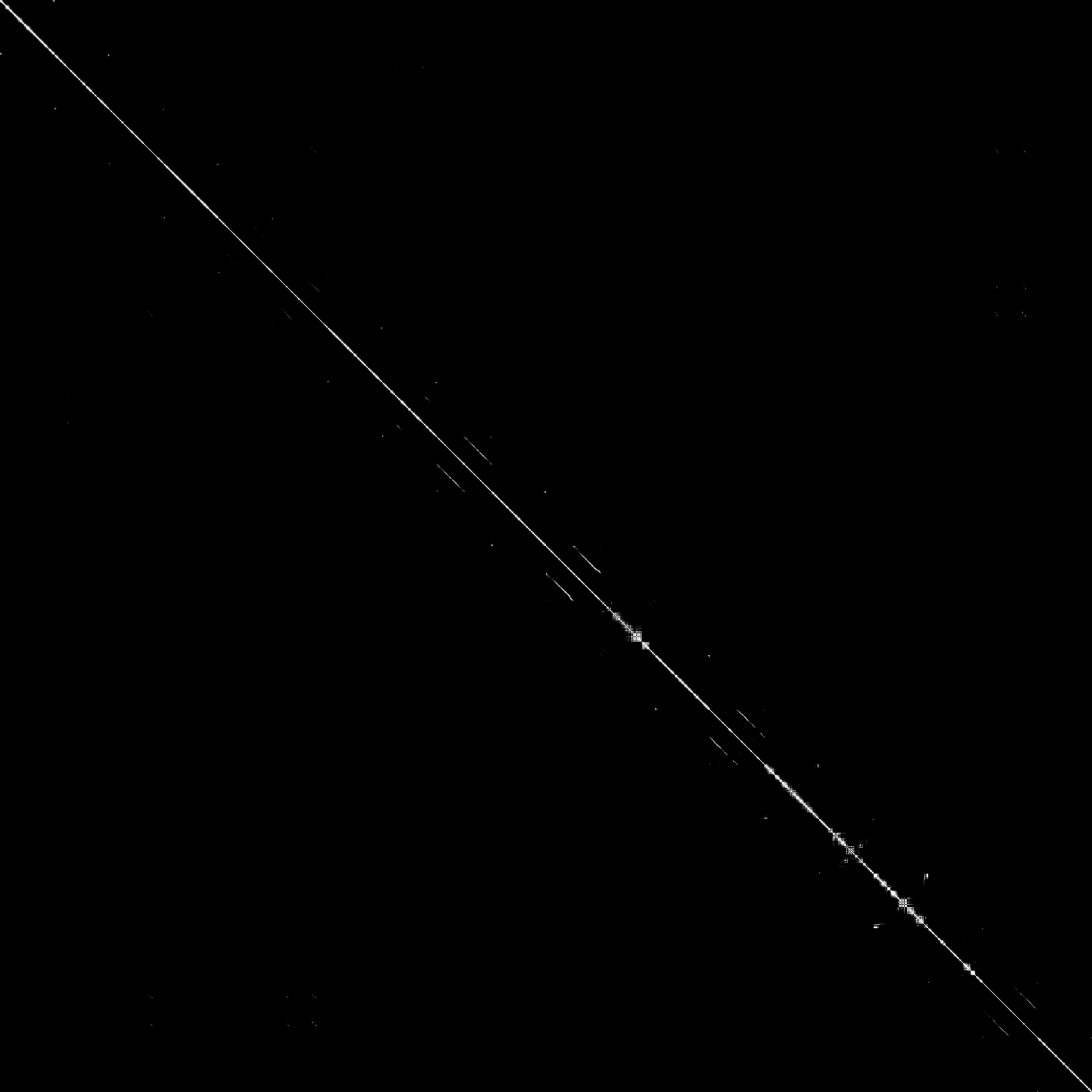}}

\subfloat[Umist-L1\label{AUmist-L1}]{\includegraphics[width=.3\textwidth]{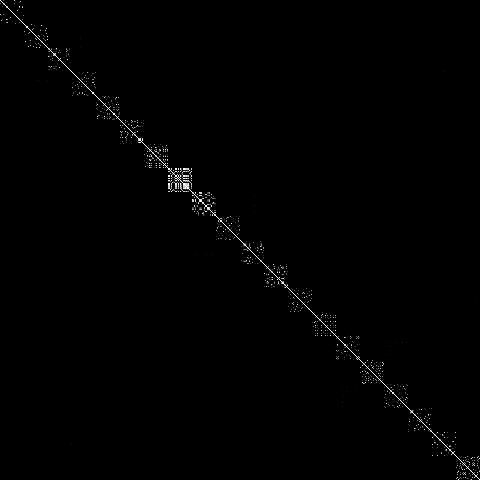}}
\hspace{.05cm}
\subfloat[Umist-L2\label{AUmist-L2}]{\includegraphics[width=.3\textwidth]{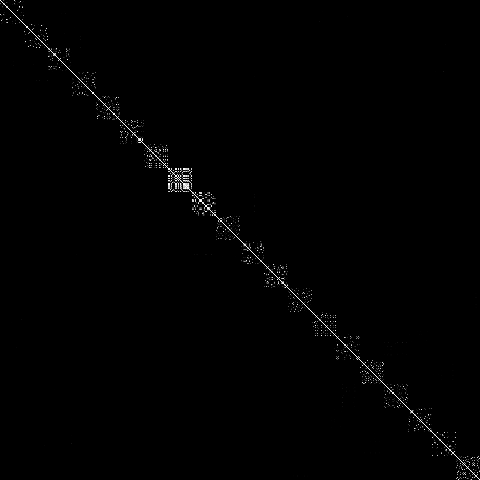}}

\subfloat[ORL-L1\label{AORL-L1}]{\includegraphics[width=.3\textwidth]{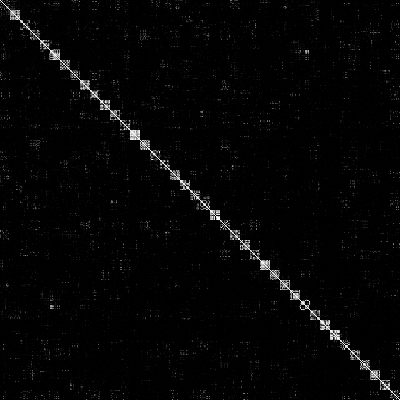}}
\hspace{.05cm}
\subfloat[ORL-L2\label{AORL-L2}]{\includegraphics[width=.3\textwidth]{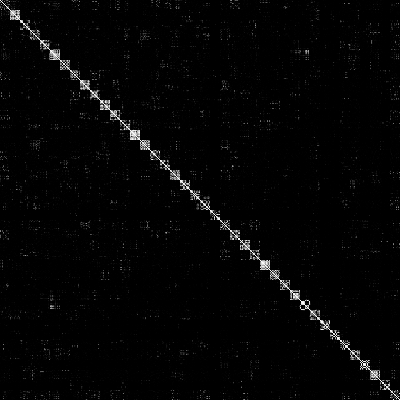}}
\caption{Visualization of learned affinity matrix $A$ on the datasets.}
\label{SimilarityA}
\end{figure}

\subsection{Results}
Clustering results based on similarity learning is recorded in Table \ref{cluster-er}. As observed, our method achieves the best performance among these related methods in most cases. Besides, we have these observations:

\begin{itemize}
\item{RGRL outperforms shallow subspace clustering methods significantly. This mainly attributes to the powerful representation ability of neural networks. }
\item{RGRL significantly improves performance compared to other deep clustering methods. It benefits from the supervision of sample relations.}
\item{RGRL performs better than RGRL$_{sc}$ on all datasets except EYaleB. For example, in terms of accuracy, RGRL outperforms RGRL$_{sc}$ by about 6\% on MNIST and Umist. This verifies the importance of locality preserving.
}
\item{With respect to deep subspace clustering, the improvement is also impressive. For instance, accuracy improves by 8\% on Umist. Moreover, DASC \cite{zhou2018deep} and S$^2$ConvSCN \cite{zhang2019self} are also two related methods. Since they have not released their code, we directly cite their results on the datasets we both used. On COIL20/COIL40, our acc is 0.9701/0.8396, while DASC gives 0.9639/0.8354. On MNIST, our acc is 0.8140, while DASC gives 0.8040. On Umist, our acc is 0.8104, while DASC gives 0.7688. On EYaleB, ours is 0.984, while DASC gives 0.9856, S$^2$ConvSCN gives 0.9848. On ORL, ours is 0.87, while DASC gives 0.8825, S$^2$ConvSCN gives 0.895. It proves that our proposed method is comparable or even better in some cases w.r.t. them. Note that, they come with some complicated components, such as adversarial learning and label supervision. By contrast, our proposed framework is very simple and can also incorporate those modules to further enhance the performance.}
\item{DEC, DKM, and DCCM perform even worse than shallow approaches. This is due to the fact that they use Euclidean distance or cosine distance to evaluate pairwise relation, which fails to capture the complex manifold structure. In general, subspace learning approach works much better in this situation. Compared to our method, DEPICT \cite{ghasedi2017deep} and DSCDAN \cite{yang2019deep} produce inferior performance. With respect to DEPICT, the performance of DSCDAN is more stable.}

\end{itemize}

To intuitively show the merit of our similarity learning approach, we visualize the affinity matrix $A$ in Fig. \ref{SimilarityA}, where $A_{ij}$ indicates the similarity between $X_i$ and $X_j$ and brighter pixel means higher similarity. Since the indexes of samples are sorted by classes, the ideal matrix should have a block-diagonal structure.
It can be seen that the similarity matrix $A$ learned by our algorithm well exhibits this block-diagonal property and it is hard to detect much difference between $\ell_\text{1}$ and $\ell_\text{2}$-norm. In particular, for the MNIST data set, the block size is relatively large since each cluster contains the most points. It can be observed that the block is very obvious, which indicates that the energy is more evenly distributed within each class. For the Umist dataset, it contains 20 classes and each single block only consists of 24 samples. The ORL data set contains 40 classes and each block only contains 10 instances. Therefore, it is a very challenging data. By observing Fig. \ref{AORL-L1} and \ref{AORL-L2}, we can see that the energy distribution still satisfies the block-diagonal property and the similarity within each class is relatively uniform, which guarantees a good performance.

It is worth noting that Figs. \ref{ACOIL20-L1} and \ref{ACOIL20-L2} are quite different from others. More concretely, the energy of similarity graph is mainly distributed on the diagonal and its very small neighborhood, which means that each sample only has a strong connection with few nearby points. This phenomenon could possibly be explained by the acquisition process of COIL20. COIL20 has 20 kinds of objects and each picture is obtained by taking a photo every 5 degrees of rotation. As a result, each picture has a very strong connection to the photos with similar angles, especially the two that are taken before and after it. In addition, notice that some small dots reside on a line parallel to the diagonal line. They correspond to the first and the last sample in each class, which are supposed to be similar. The similarity between different classes is still small, so that each class can be finally separated. 
\begin{figure}[!htpb]
\centering
\subfloat[MNIST-L1]{\includegraphics[width=.4\textwidth]{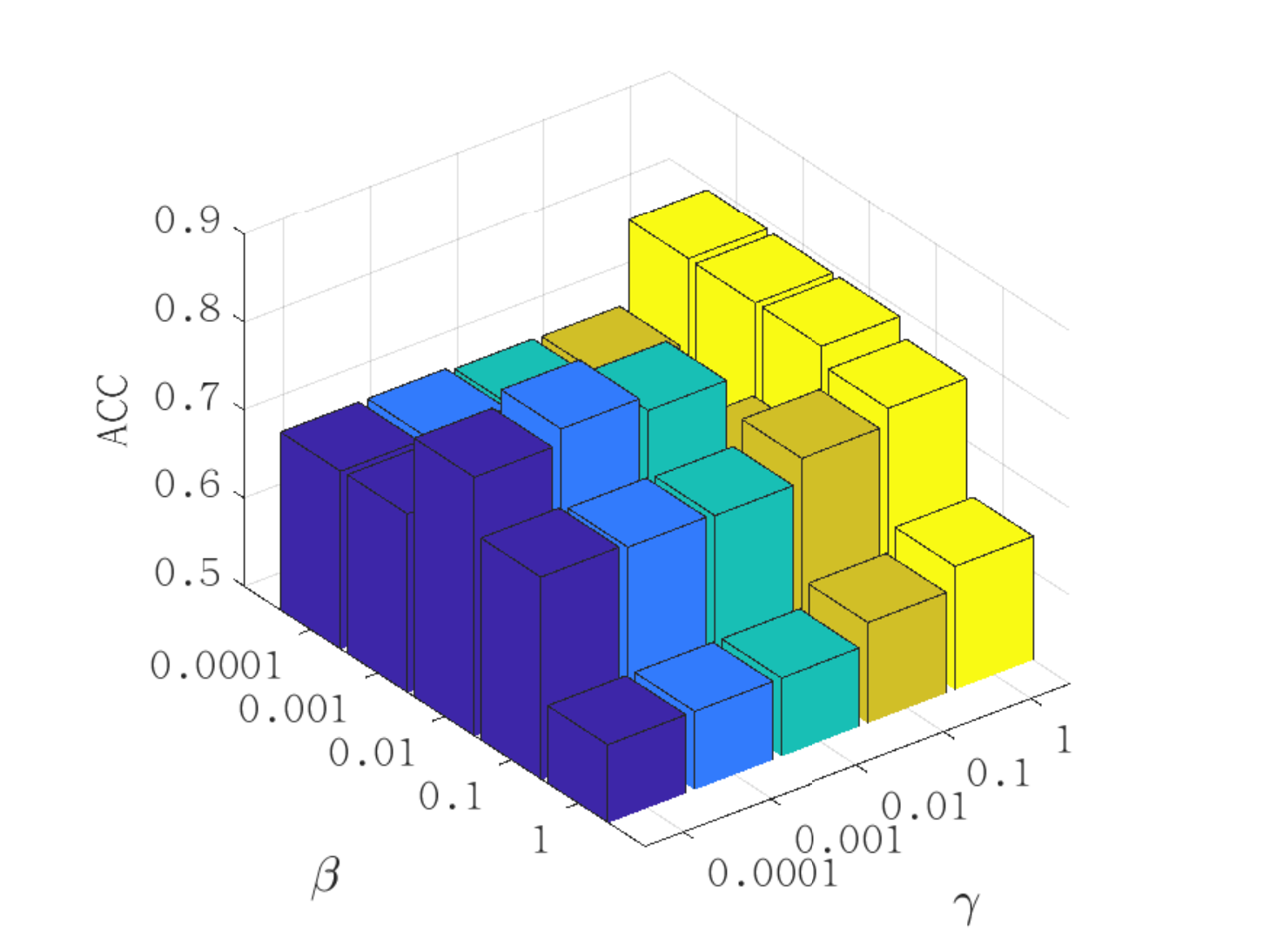}}
\vspace{-.4cm}
\subfloat[MNIST-L2]{\includegraphics[width=.4\textwidth]{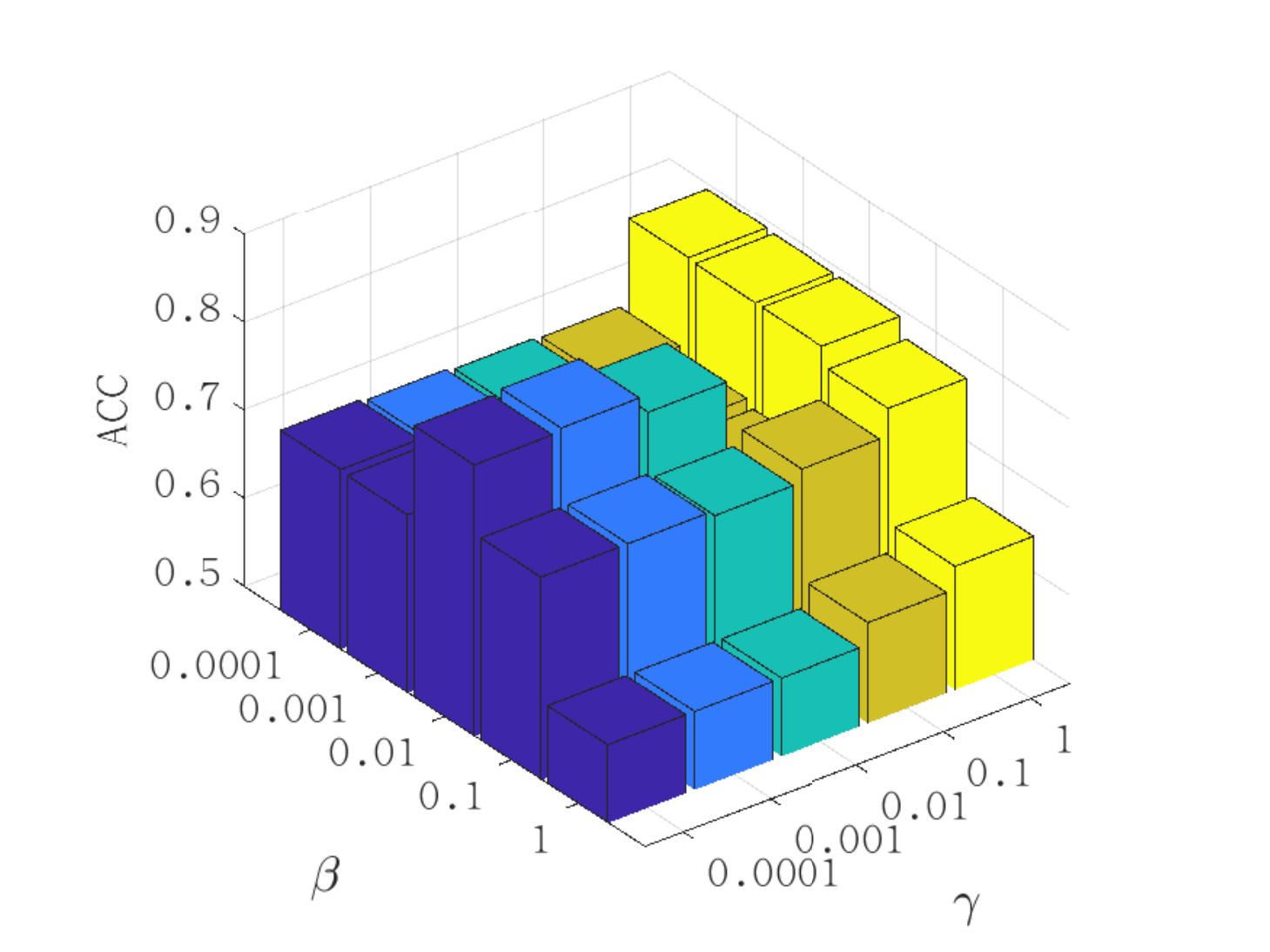}}

\subfloat[COIL20-L1]{\includegraphics[width=.4\textwidth]{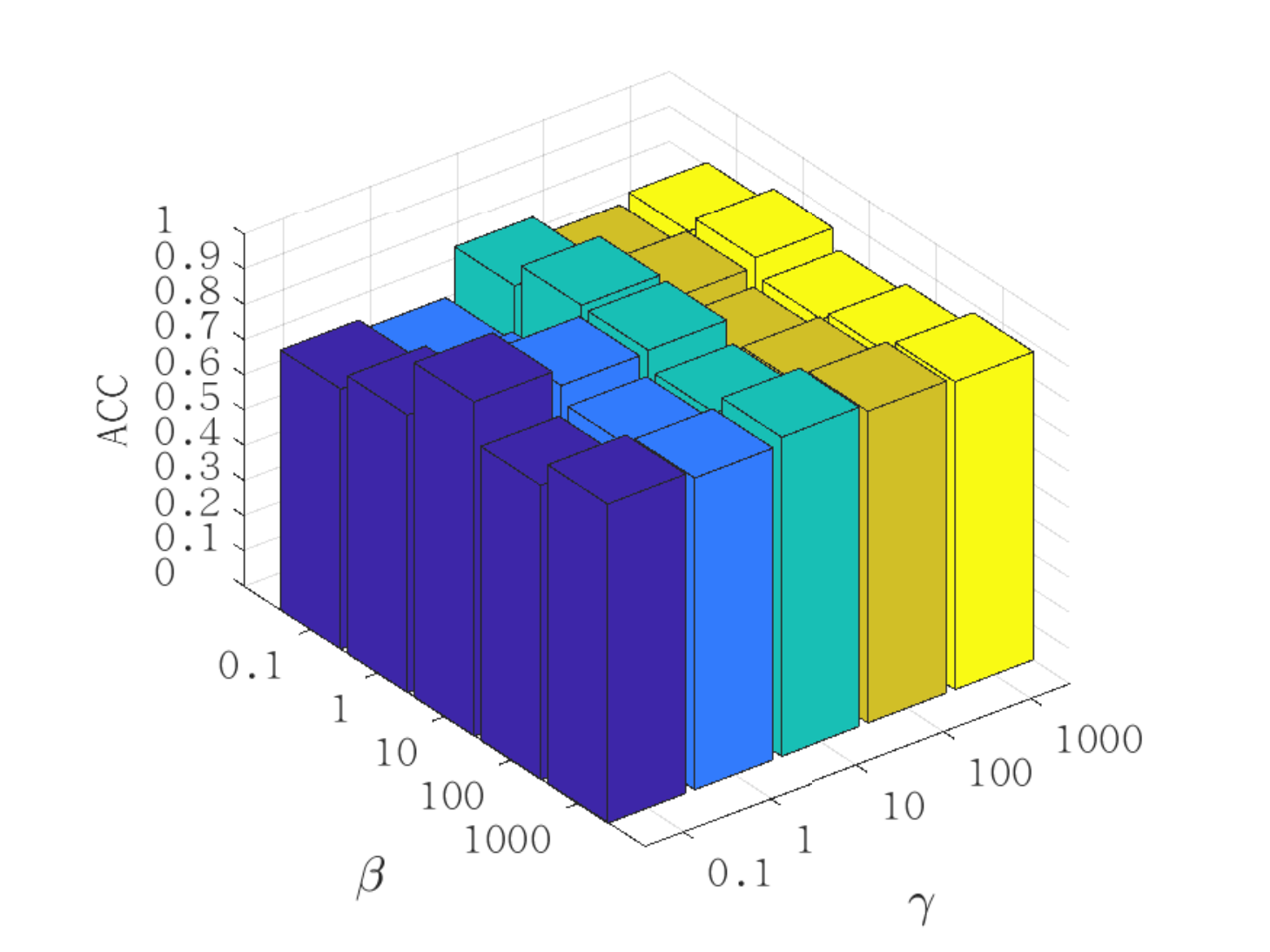}}
\vspace{-.4cm}
\subfloat[COIL20-L2]{\includegraphics[width=.4\textwidth]{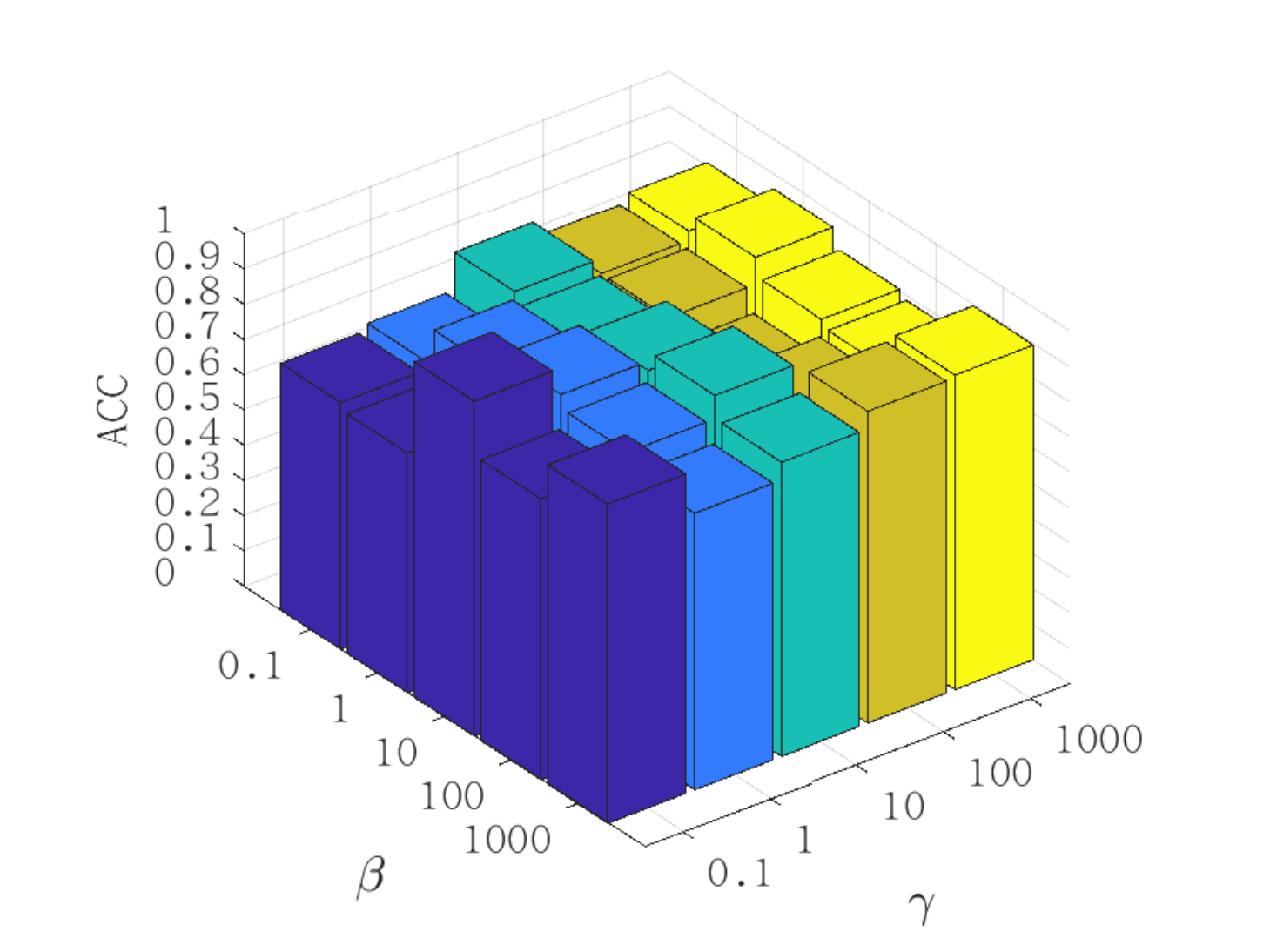}}

\subfloat[Umist-L1]{\includegraphics[width=.4\textwidth]{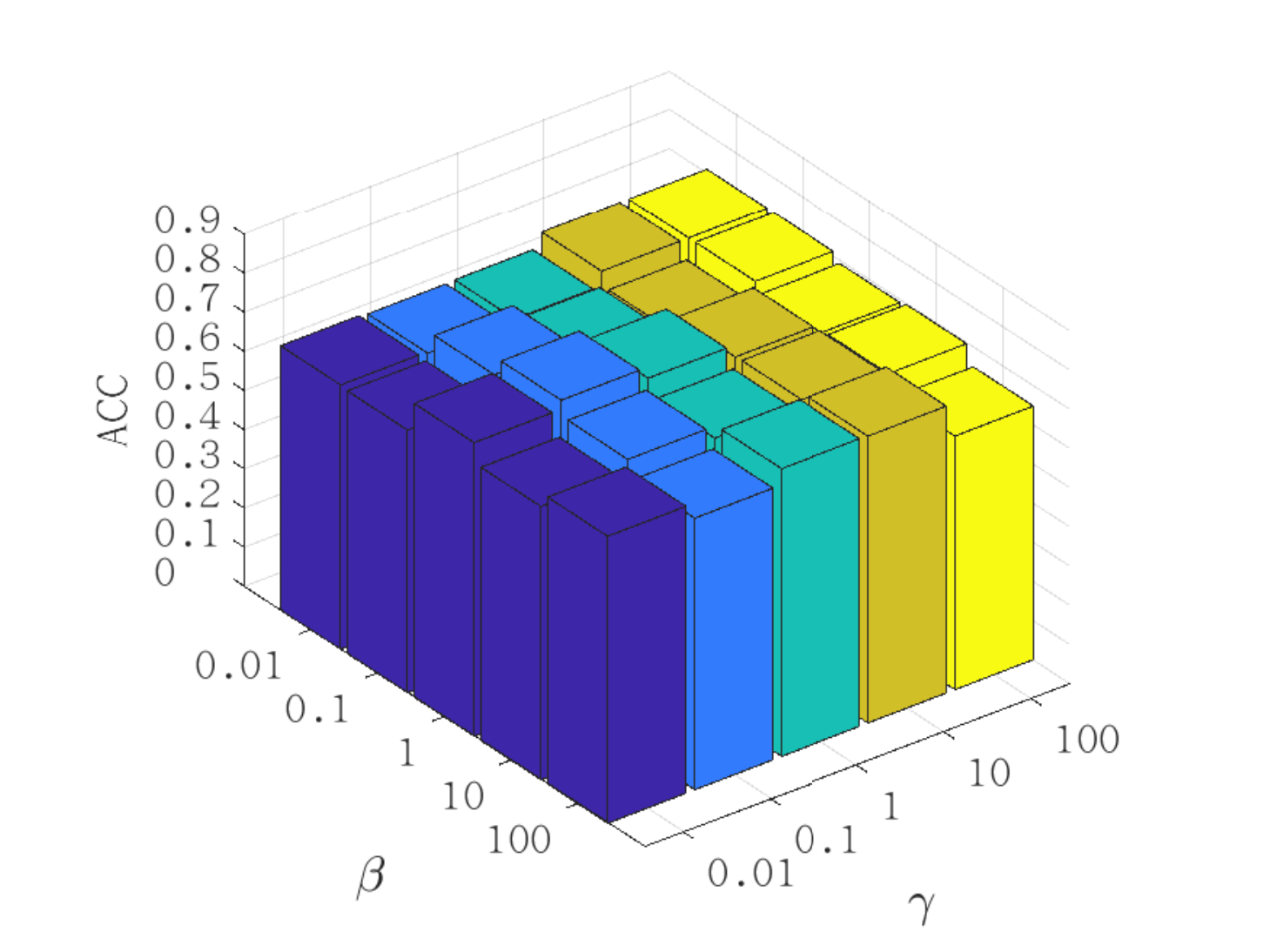}}
\vspace{-.4cm}
\subfloat[Umist-L2]{\includegraphics[width=.4\textwidth]{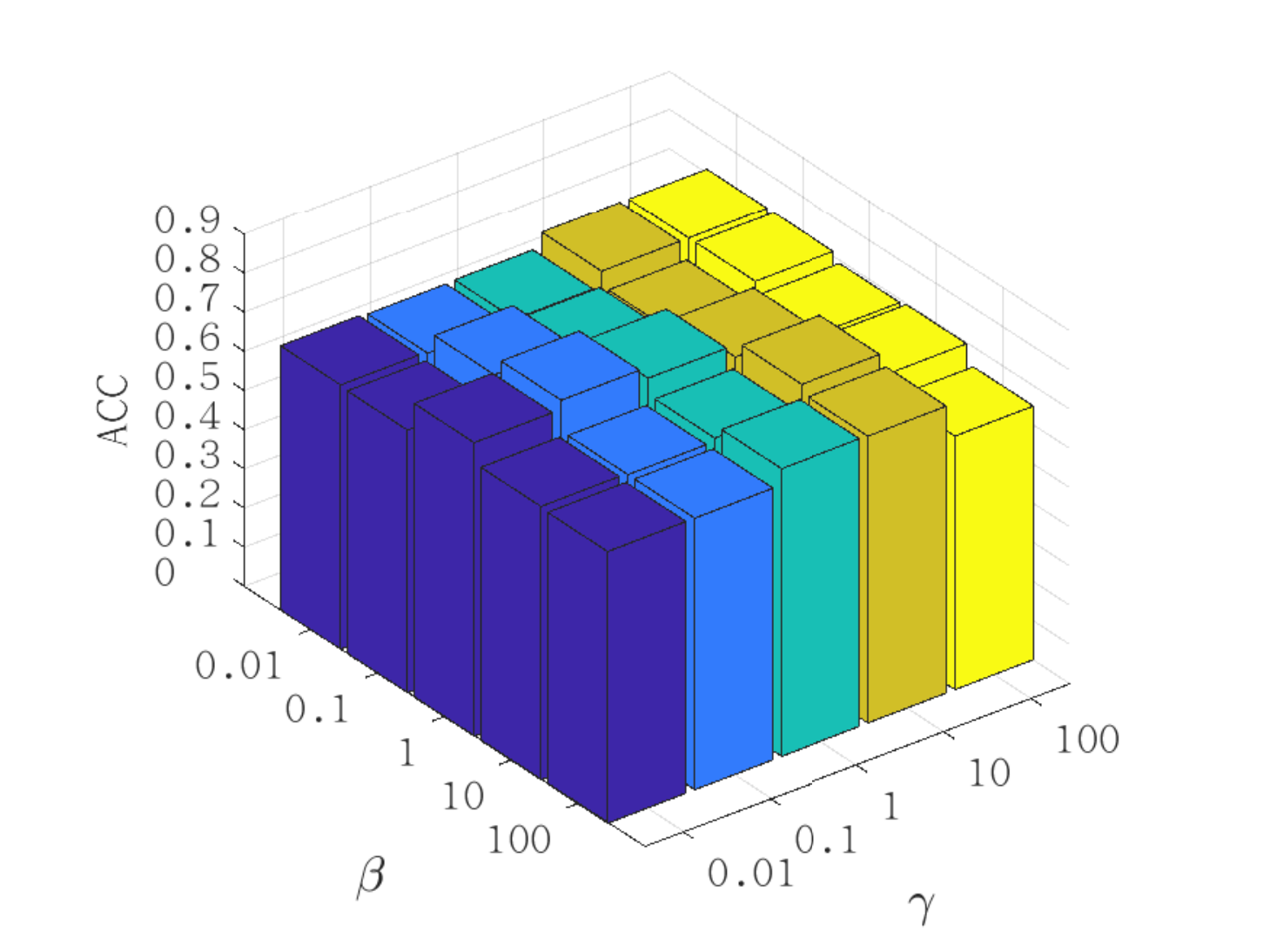}}

\subfloat[ORL-L1]{\includegraphics[width=.4\textwidth]{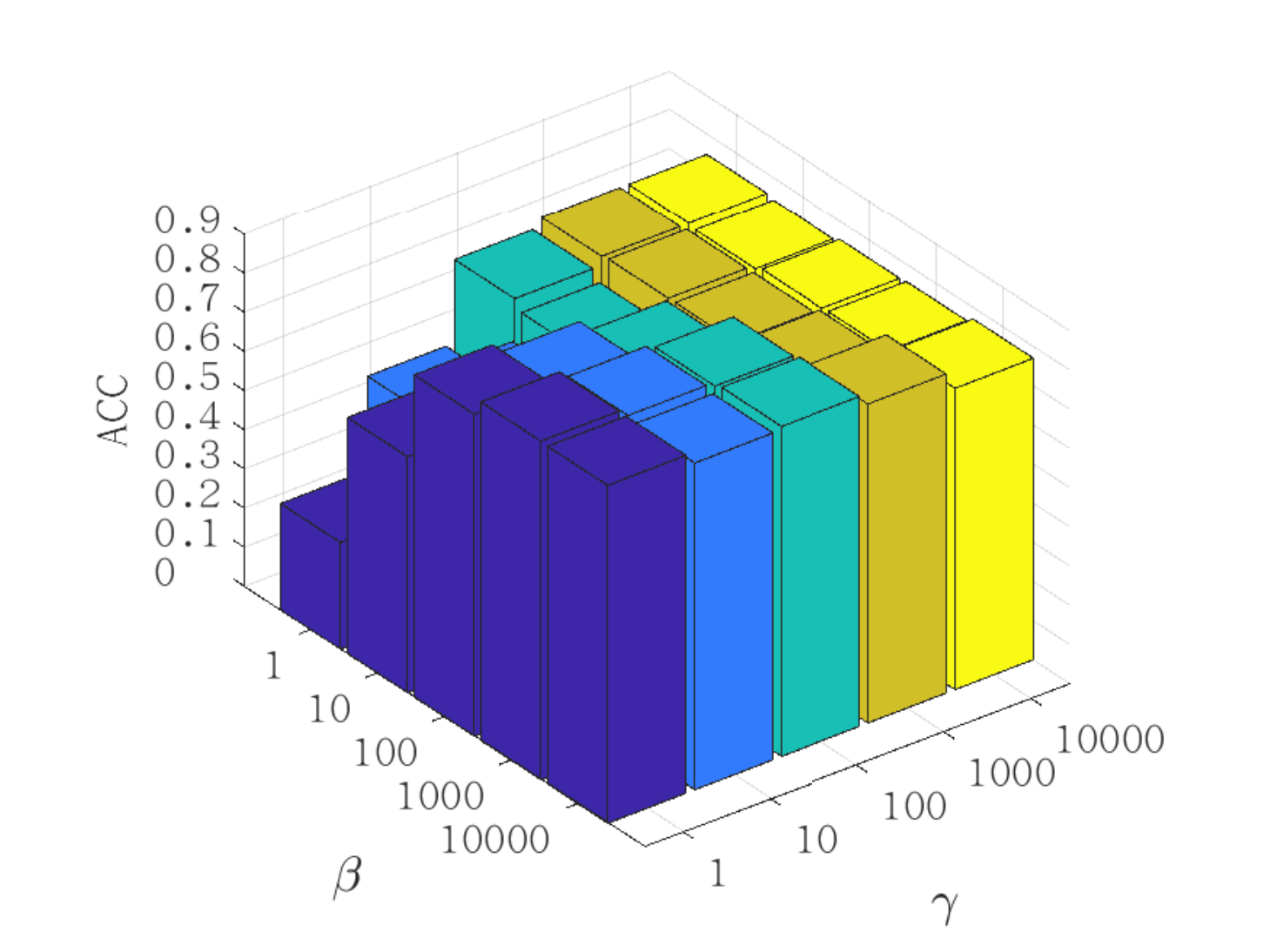}}
\vspace{-.4cm}
\subfloat[ORL-L2]{\includegraphics[width=.4\textwidth]{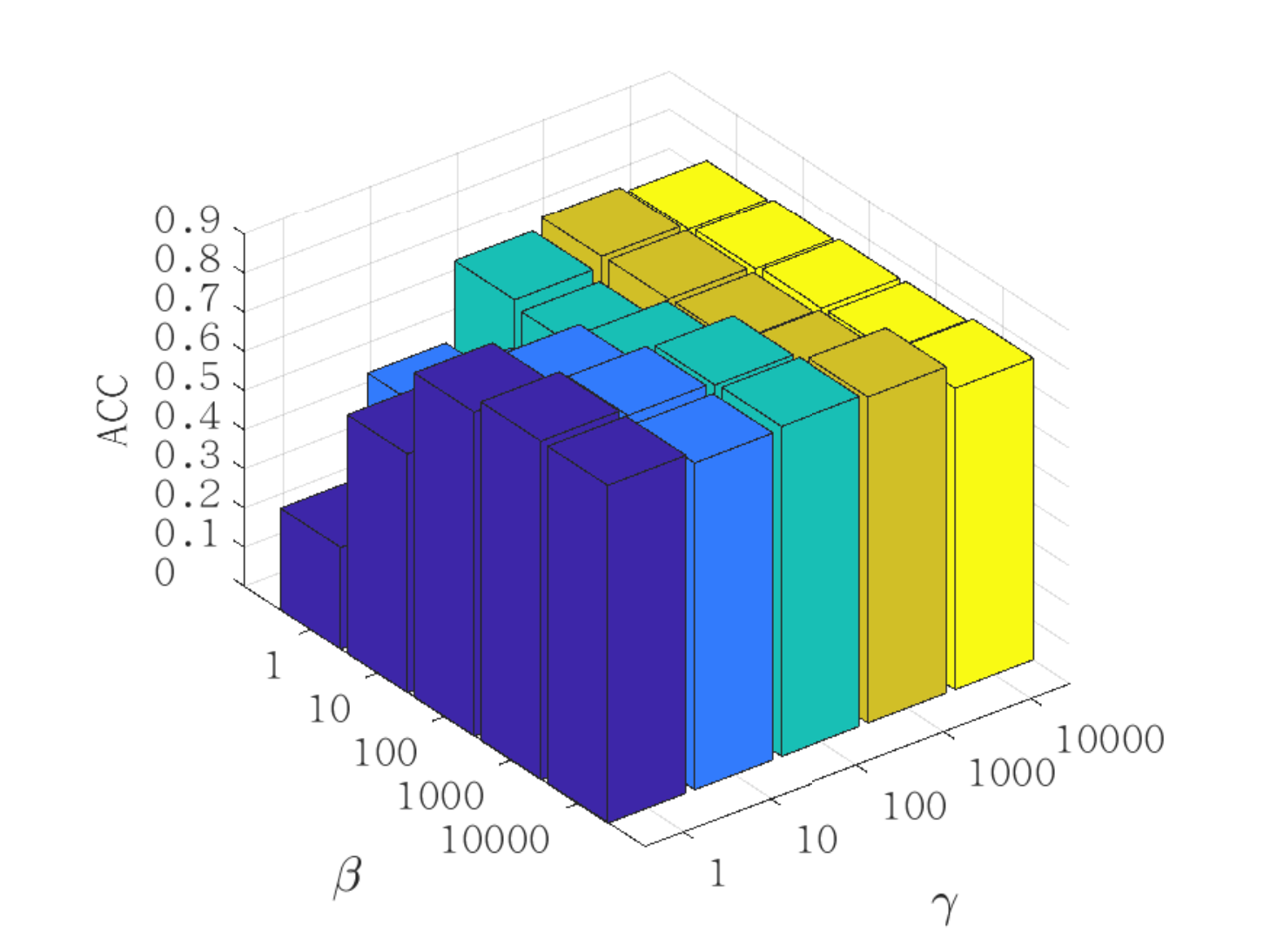}}
\caption{The influence of parameters on accuracy of the datasets.}
\label{mnist-acc}
\end{figure}
\subsection{Parameter Analysis}
There are three hyper-parameters in our model: $\alpha$, $\beta$ and $\gamma$. As mentioned earlier, $\alpha$ is fixed as 1e-4. We show the variation of accuracy along with the change of $\beta$ and $\gamma$ in Fig. \ref{mnist-acc}. Generally speaking, parameters are data-specific. Thus, different ranges are searched for different data sets. It can be seen that the performance on MNIST is very sensitive to the value of $\beta$, which means that subspace consistence plays a crucial role. For other datasets, our method works well for a wide range of values. One possible reason for the fluctuations is due to the fact that spectral clustering method is sensitive to the graph. It is well-known that a small disturbance in graph can lead to a large difference in clustering performance. 

\begin{table}
\centering
\caption{Statistics of the large datasets.}
\renewcommand{\arraystretch}{1.1}
\resizebox{.45\textwidth}{!}{
\begin{tabular}{c|c c c c}
\hline
Dataset & MNIST & USPS & RCV1 \\
\hline
Samples & 70,000 & 9,298 & 10,000 \\
\hline
Classes & 10 & 10 & 4 \\
\hline
Dimensions & 28$\times$ 28 & 16$\times$ 16 & 2,000\\
\hline
\end{tabular}}
\label{dataset-large}
\end{table}

\begin{table*}[ht]
\centering
\caption{Clustering results on MNIST, USPS, and RCV1.}
\renewcommand{\arraystretch}{1.1}
\resizebox{.9\textwidth}{!}{
\begin{tabular}{c |c|c c c c c c c c c c}
\hline
Dataset & Metric & KM & AE+KM & DCN & IDEC & DKM & DCCM & DEPICT & DSCDAN & RGRL-L1 & RGRL-L2 \\
\hline
 & ACC & 0.535 & 0.808 & 0.811 & 0.857 & 0.840 & 0.655 & \textbf{0.9295} & 0.8189 & {0.9127} & {0.9127}\\
MNIST & NMI & 0.498 & 0.752 & 0.757 & {0.864} & 0.796 & 0.679 & \textbf{0.8799} & 0.8727 & 0.8175 & 0.8175\\
 \hline
& ACC & 0.673 & 0.729 & 0.730 & 0.752 & 0.757 & 0.686 & 0.8565 & 0.8061 & 0.9148 & \textbf{0.9170}\\
USPS & NMI & 0.614 & 0.717 & 0.719 & 0.749 & 0.776 & 0.675 & 0.8652 & 0.8507 & \textbf{0.8449} & 0.8320\\
\hline
& ACC & 0.508 & 0.567 & 0.567 & 0.595 &  0.583 & - & - & - & 0.6852 & \textbf{0.6867}\\
RCV1 & NMI & 0.313 & 0.315 & 0.316 & 0.347 &  0.331 & - & - & -  & 0.4019 & \textbf{0.4030}\\
\hline
\end{tabular}}
\label{cluster-big}
\end{table*}

\section{Out-of-sample Experiments}
\label{secemb-exp}
In this section, we further show that our method can be extended to address large-scale and out-of-sample problem by embedding samples into subspace. 
\subsection{Datasets}
The datasets used in this experiment are large clustering collections, each is composed of more than 9,000 samples. We perform experiments on two object datasets: MNIST and USPS; a text dataset: RCV1. We use full MNIST dataset which has 70,000 hand-written digits images in 10 classes. USPS contains 9,298 hand-written digits images in 10 classes. RCV1 contains around 810,000 English news stories labeled with a category tree. Following DKM \cite{fard2018deep}, we randomly sample 10,000 documents from the four largest categories: corporate/industrial, government/social, markets and economics of RCV1, each sample only belongs to one of these four categories. Note that, different from experiments in DEC \cite{xie2016unsupervised} and IDEC \cite{guo2017improved}, we keep samples with multiple labels only if they don't belong to any two of selected four categories. For text datasets, we select 2000 words with the highest tf-idf values to represent each document. The statistics of datasets in this experiment are summarized in Table \ref{dataset-large}.

\subsection{Experimental Setup}
Different from the above experiment, we use latent representations for clustering task. In particular, we train the network with a reasonable small batch of samples (5,000 samples for each dataset in our experiments), then we use the similarity matrix to predict the pseudo-labels of selected samples by the above approach. Finally, we encode all the data into latent space and use a nearest-neighbor classifier to predict the labels for the rest of the data. Following this approach, our method can address out-of-sample problem.

For fair of comparison, we use the same encoder/decoder architecture as DEC \cite{xie2016unsupervised}, IDEC \cite{guo2017improved}, and DKM \cite{fard2018deep}. The encoder is a fully connected network with dimensions of $d$-500-500-2000-$k$ for all datasets, where $d$ is the dimension of input features and $k$ is the number of clusters. And the decoder correspondingly is a mirror of the encoder, a fully connected network with dimensions of $k$-2000-500-500-$d$. A ReLU activate function is applied for each layer except the input, output, and embedding layer. We pre-train the auto-encoder 50 epochs and fine-tune the whole network with objective function (\ref{objf}) 30 epochs.

We compare our method with the $k$-means clustering (KM), an auto-encoder followed by $k$-means applied to the latent representation (AE+KM), and recent deep clustering approaches: deep clustering network (DCN) \cite{yang2017towards}, IDEC \cite{guo2017improved}, DKM \cite{fard2018deep}, DCCM \cite{wu2019deep}, DEPICT \cite{ghasedi2017deep}, and DSCDAN \cite{yang2019deep}. Since DCCM, DEPICT, and DSCDAN are designed for image dataset, they can not apply to text data RCV1.

\begin{figure*}[ht]
\centering
\subfloat[IDEC\label{IDEC-v}]{\includegraphics[width=.3\textwidth]{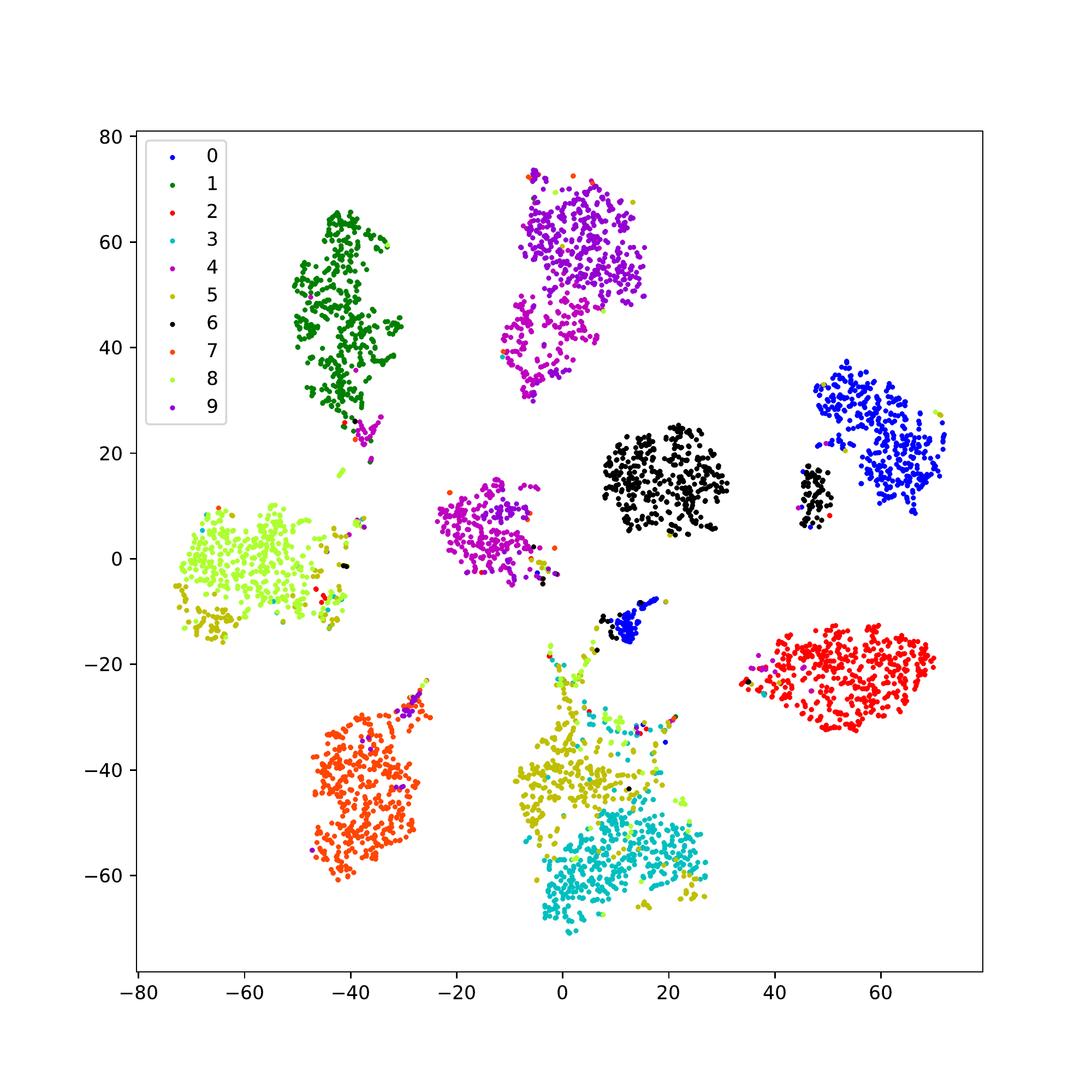}}
\subfloat[DKM\label{DKM-v}]{\includegraphics[width=.3\textwidth]{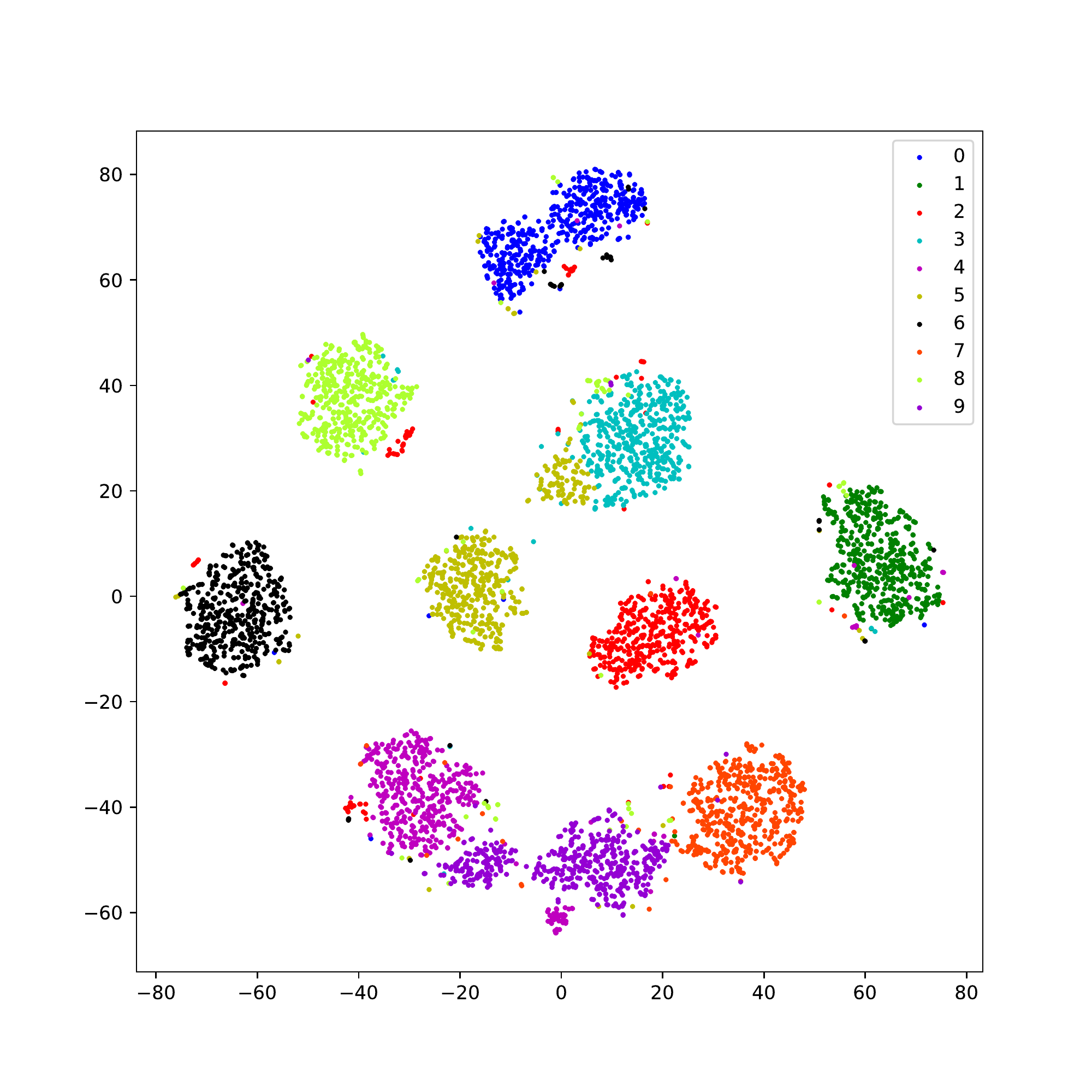}}
\subfloat[DCCM\label{DCCM-v}]{\includegraphics[width=.3\textwidth]{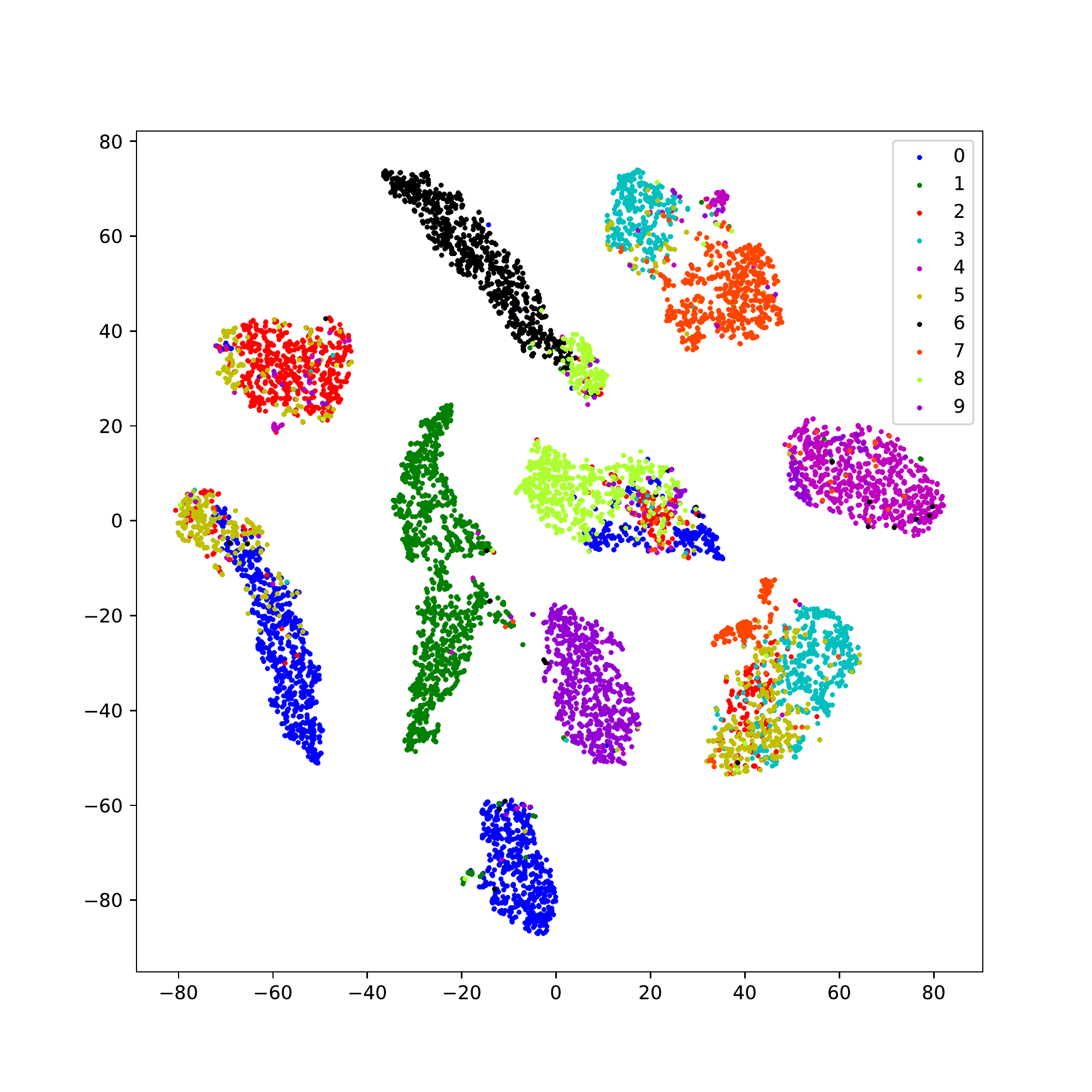}}\\
\subfloat[DEPICT\label{DEPICT-v}]{\includegraphics[width=.3\textwidth]{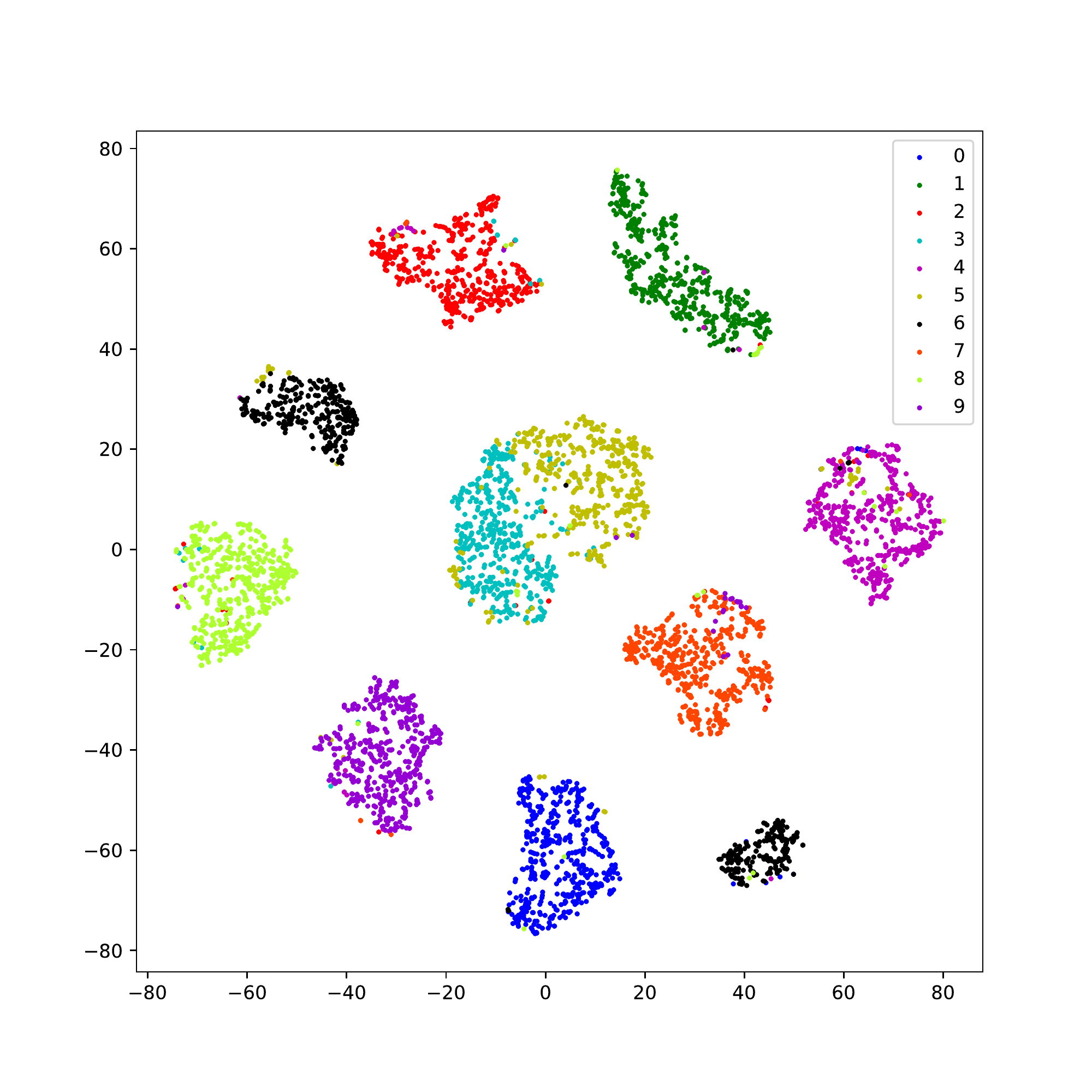}}
\subfloat[DSCDAN\label{DSCDAN-v}]{\includegraphics[width=.3\textwidth]{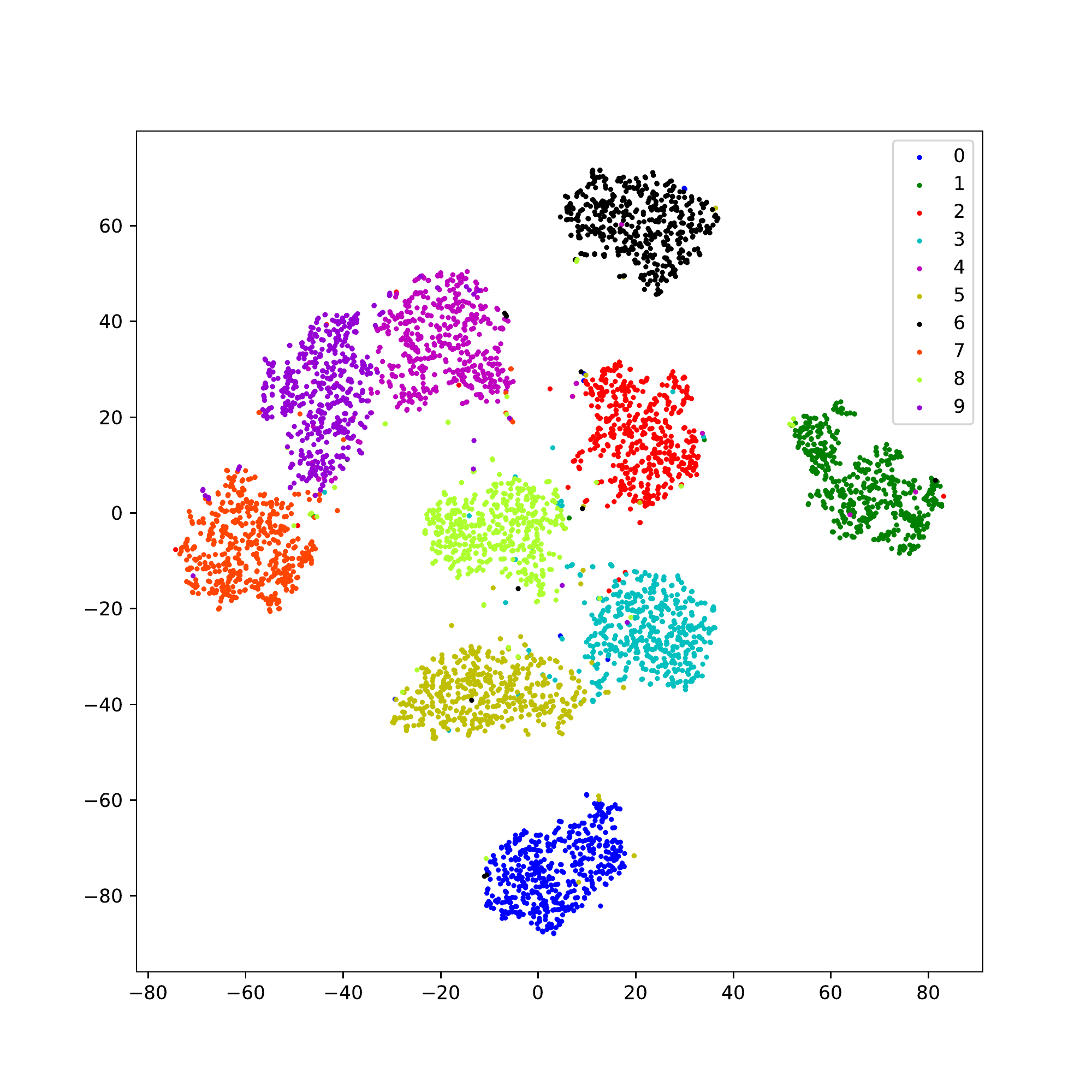}}
\subfloat[RGRL-L2\label{SAE-v}]{\includegraphics[width=.3\textwidth]{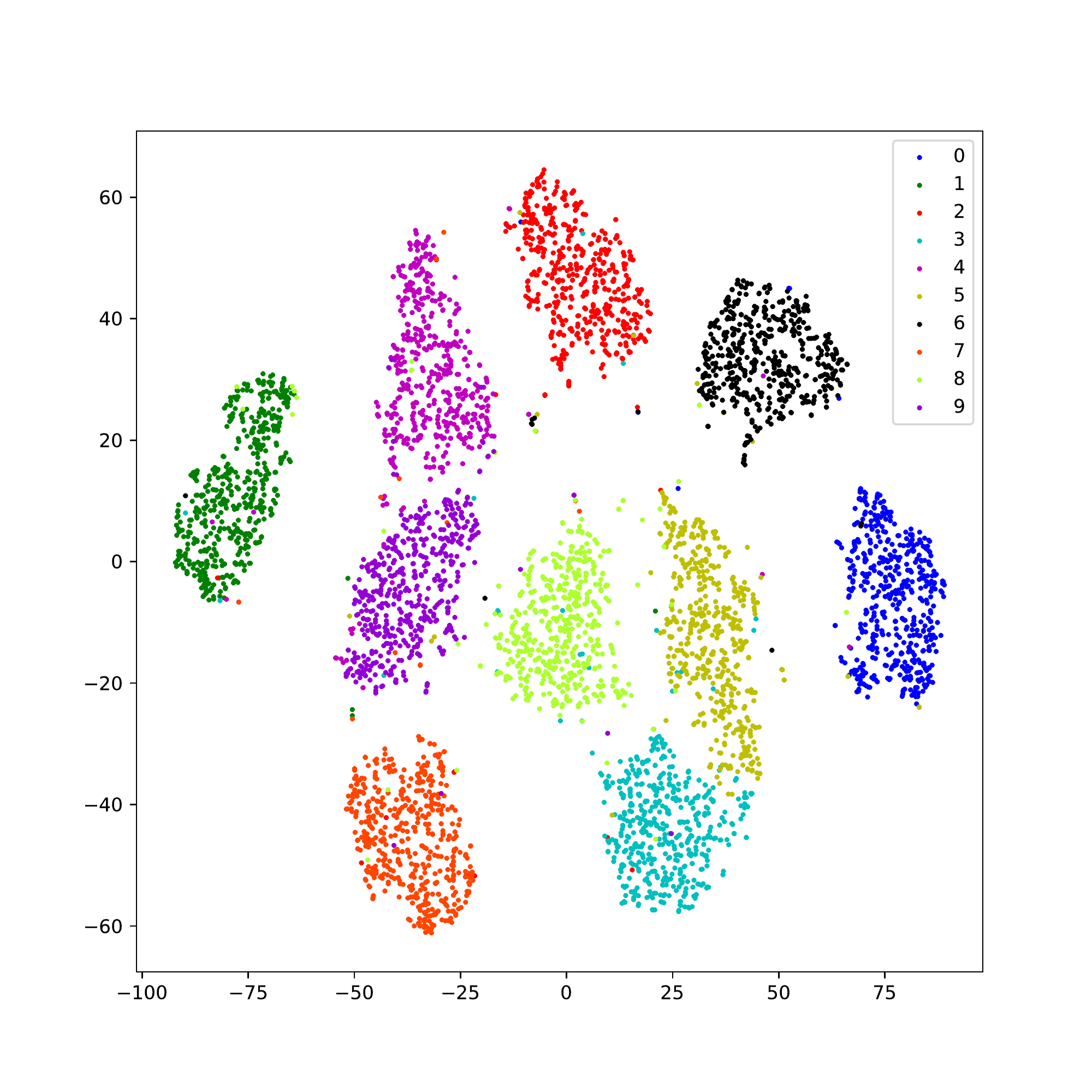}}
\caption{2D visualization of the embedding spaces learned on USPS dataset.}
\label{tsne}
\end{figure*}

\subsection{Results}
Clustering performance of embedding experiment is recorded in Table \ref{cluster-big}. Our method still outperforms recent deep clustering methods. In particular, compared to recent DKM, RGRL-L2 improves accuracy by 7.27\%, 16\%, and 10.37\% on those three datasets, respectively. With regard to DCCM and DSCDAN, the accuracy improvement is more than 20\% and 10\%, respectively. The accuracy of our method is a little bit lower than DEPICT on MNIST, but it is improved by 6\% on USPS.  These benefits from embedding samples into subspaces. Therefore, our method proved to be an attractive technique to deal with large scale and out-of-sample problem. 

We use the t-SNE method to visualize the learned latent representations of four most recent methods. As we can observe from Fig. \ref{tsne}, samples of different classes tangle in IDEC and DKM, which is because they force samples to move to cluster centers. Same phenomenon happens on DCCM since it is supervised by pseudo-graph guided by cosine similarity. For DEPICT, the black points are located in two separate places, which
could degrade the performance. For DSCDSN, we can see that several clusters are close to each other, which will deteriorate the performance. Our method aims to project samples of each class into a subspace, thus samples can be well separated. 

\section{Conclusion}
\label{secconclusion}
In this paper, we have presented a novel representation learning network, which is guided by sample relations learned by the network itself. To the best of our knowledge, it is the first effort to preserve both local neighborhood and global subspace consistency. Extensive experimental results on both small scale and large scale datasets have shown the superiority of the proposed method on similarity and representation learning over state-of-the-arts, including the latest deep learning based method. Note that our model is simple and fundamental, more complicated components, such as adversarial learning, label supervision, can be added to our framework to further improve the performance.
\section{Acknowledgement}
This paper was in part supported by Grants from the Natural
Science Foundation of China (No. 61806045), the National Key R\&D Program of China (No. 2018YFC0807500), the Fundamental Research Fund for the Central Universities under Project ZYGX2019Z015, the Sichuan Science and Techology Program (Nos. 2020YFS0057, 2019YFG0202), the Ministry of Science and Technology of Sichuan Province Program (Nos. 2018GZDZX0048, 20ZDYF0343, 2018GZDZX0014, 2018GZDZX0034).
\section{References}
\bibliographystyle{elsarticle-num}
\bibliography{ref}

\end{document}